\pdfoutput=1
\documentclass[11pt]{article}

\usepackage[utf8]{inputenc}
\usepackage[T1]{fontenc}
\usepackage{booktabs}
\usepackage{wrapfig}
\usepackage{longtable}
\usepackage{xurl}
\usepackage{amsfonts}
\usepackage{nicefrac}
\usepackage{microtype}
\usepackage[table]{xcolor}
\PassOptionsToPackage{hyphens}{url}

\usepackage{smile}
\usepackage{tikz}
\usetikzlibrary{matrix, positioning}
\usepackage{pgfplots}
\pgfplotsset{compat=1.18}
\usepackage{wrapfig}
\usepackage{needspace}
\usepackage{xcolor}
\usepackage{cancel}
\usepackage{acronym}

\definecolor{mypink}{HTML}{FF1493}
\definecolor{darkred}{HTML}{BB253B}
\definecolor{lightgreen}{HTML}{90C0BF}

\definecolor{darkcyan}{HTML}{0081B9}

\newmdenv[
  backgroundcolor=gray!10,
  skipabove=1em,
  skipbelow=1em,
  leftline=false,
  topline=false,
  bottomline=false,
  rightline=false,
  linecolor=gray!88,
  linewidth=4pt
]{githubquote}

\usepackage[most,skins,theorems]{tcolorbox}
\tcbset{
  aibox/.style={
    width=\linewidth,
    top=10pt,
    bottom=4pt,
    colback=blue!6!white,
    colframe=black,
    colbacktitle=black,
    enhanced,
    center,
    attach boxed title to top left={yshift=-0.1in,xshift=0.15in},
    boxed title style={boxrule=0pt,colframe=white,},
  }
}

\definecolor{rliableolive}{HTML}{BBCC33}
\definecolor{rliableblue}{HTML}{77AADD}
\definecolor{rliablered}{HTML}{EE8866}
\definecolor{LightCyan}{rgb}{0.88,1,1}
\definecolor{darkblue}{HTML}{2878D9}
\definecolor{navyblue}{HTML}{0000FF}

\newtcolorbox{AIbox}[2][]{aibox,title=#2,colback=rliableblue!10!white,#1}

\providecolor{yaleblue}{HTML}{00356B}
\colorlet{findingframe}{yaleblue}
\definecolor{findingaccent}{HTML}{D97757}
\colorlet{findingbg}{yaleblue!5!white}

\newtcolorbox[%
  auto counter,
  number within=section,
  crefname={finding}{findings},
  Crefname={Finding}{Findings},
]{finding}[1][]{%
  enhanced,
  breakable,
  before skip=10pt,
  after skip=10pt,
  colback=findingbg,
  colframe=findingframe,
  boxrule=0pt,
  arc=3pt,
  outer arc=3pt,
  left=12pt, right=12pt, top=8pt, bottom=10pt,
  before upper={%
    \noindent\textbf{Finding~\thetcbcounter\ifblank{#1}{}{. #1}.}\enspace
  },
}

\usepackage{subcaption}

\usepackage{titlesec}
\usepackage{xcolor}
\definecolor{sectionblue}{HTML}{00356B}

\titleformat{\section}
  {\Large\bfseries\color{sectionblue}}{\thesection}{1em}{}
\titleformat{\subsection}
  {\large\bfseries\color{sectionblue}}{\thesubsection}{1em}{}
\titleformat{\subsubsection}
  {\normalsize\bfseries\color{sectionblue}}{\thesubsubsection}{1em}{}
\titleformat{\paragraph}[runin]
  {\normalfont\normalsize\bfseries\color{black}}{}{0pt}{}

\usepackage[palantir]{paperstyle}

\hypersetup{hypertexnames=false}
\renewcommand*{\backref}[1]{}

\graphicspath{{figs/}}
\usetikzlibrary{arrows.meta}
\definecolor{laBlueLine}{HTML}{5C81AD}
\definecolor{laOrangeLine}{HTML}{C18C54}
\definecolor{laPurpleLine}{HTML}{9684B0}
\NewDocumentCommand{\eqcallout}{O{laBlueLine} O{0pt} O{eqcsym} m m}{%
  \tikz[baseline=(#3.base),remember picture]{%
    \node[inner xsep=2pt,inner ysep=1.5pt,rounded corners=1.5pt,fill=#1!16] (#3) {$#5$};
    \node[overlay,anchor=base,inner xsep=3pt,inner ysep=1.5pt,text height=6pt,text depth=1.5pt,
      rounded corners=1pt,fill=#1!10,font=\sffamily\footnotesize,text=#1!85!black]
      (eqclabel) at ([xshift=#2,yshift=20pt]#3.base) {#4};
    \draw[overlay,-{Stealth[length=1.1mm]},#1!70,line width=.35pt]
      (eqclabel.south) -- ([yshift=1pt]#3.north);
  }}
\NewDocumentCommand{\eqbox}{O{laBlueLine} m m}{%
  \tikz[baseline=(#2.base),remember picture]{%
    \node[inner xsep=2pt,inner ysep=1.5pt,rounded corners=1.5pt,fill=#1!16] (#2) {$#3$};
  }}
\newif\iftermtint
\NewDocumentCommand{\termhl}{O{laBlueLine} m}{%
  \iftermtint
  \tikz[baseline=(hlsym.base)]{%
    \node[inner sep=0pt,outer sep=0pt,text height=.7em,text depth=.2em] (hlsym) {\phantom{#2}};
    \fill[overlay,#1!10,rounded corners=1pt]
      ([xshift=-1.5pt,yshift=-.6pt]hlsym.south west) rectangle ([xshift=1.5pt,yshift=.6pt]hlsym.north east);
    \node[inner sep=0pt,outer sep=0pt,anchor=base,text=#1!85!black] at (hlsym.base) {#2};
  }%
  \else#2\fi}
\definecolor{laBlueText}{HTML}{3E6CA8}
\definecolor{laOrangeText}{HTML}{8E4C12}
\NewDocumentCommand{\walpha}{e{^_}}{{\color{laBlueText}\alpha\IfValueT{#1}{^{#1}}\IfValueT{#2}{_{#2}}}}
\NewDocumentCommand{\wbeta}{e{^_}}{{\color{laOrangeText}\beta\IfValueT{#1}{^{#1}}\IfValueT{#2}{_{#2}}}}
\newcommand{\cstream}{\termhl[laBlueLine]{current stream}}
\newcommand{\fbstream}{\termhl[laOrangeLine]{sublayer feedback}}
\definecolor{checkgreen}{HTML}{1B7F3B}
\definecolor{crossred}{HTML}{C0392B}

\newcommand{\RMSNorm}{\operatorname{RMSNorm}}

\title{Attention Latent Feedback}
\papershorttitle{Attention Latent Feedback}
\author{Xuyuan~Xiong$^{1,*}$, Siyu~Chen$^{2,*}$, Mengzhao~Chen$^{1}$, Yue~Su$^{1}$, Zhuoran~Yang$^{2}$, Ping~Luo$^{1,\dagger}$, Xihui~Liu$^{1,\dagger}$}
\paperaffiliations{$^{1}$MMLab, The University of Hong Kong\qquad $^{2}$Yale University}
\papernote{$^{*}$Equal contribution: \paperemail{xiongxuyuan@connect.hku.hk}, \paperemail{siyu.chen.sc3226@yale.edu}\\
$^{\dagger}$Corresponding: \paperemail{pluo.lhi@gmail.com}, \paperemail{xihuiliu@hku.hk}}
\paperlogos{\paperlogoitem{0.68cm}{logos/hku_mmlab}\hspace{0.5cm}\paperlogoitem{1.12cm}{logos/hku}\hspace{0.5cm}\paperlogoitem{1.08cm}{logos/yale}}
\papermeta{Date}{\today}
\papermeta{Code}{\url{https://github.com/xjxyys/ALF}}

\begin{document}

\maketitle
\pagestyle{paper}

\begin{abstract}
During autoregressive decoding, the deeper representations of the previous token have already been computed, yet shallow
layers at the current token cannot read them directly because standard self-attention connects positions only at the
same depth. We introduce Attention Latent Feedback (ALF), an architecture that allows every sublayer to attend to the outputs of the 
deeper layers at the previous position. Training unrolls this cross-position recurrence into two parallel passes,
whereas decoding remains single-pass using cached previous-token states. Across mixture-of-experts models with up to 4.0B parameters, ALF lowers held-out loss by 0.022--0.024 nats over the standard residual with about
1.1\% decoding FLOP overhead, and consistently outperforms prior final-state feedback across all three scales
on both standard and attention residual backbones. On MoE-1.6B, ALF matches a 1.38$\times$ larger baseline
token budget, and on MoE-4.0B it lifts average accuracy across seventeen downstream benchmarks from 45.1\% to
47.0\%, with consistent gains under single-pass autoregressive generation. Analysis further reveals that intermediate and nearby deeper layers drive most of this benefit,
allowing a fixed nearest-window variant to retain 89\%--94\% of the gain at linear read
cost while learned attention widens its margin with scale.
\end{abstract}
\suppressfloats[t]
\enlargethispage{12pt}
{\setlength{\intextsep}{6pt}
\begin{figure}[!ht]
  \centering
  \includegraphics{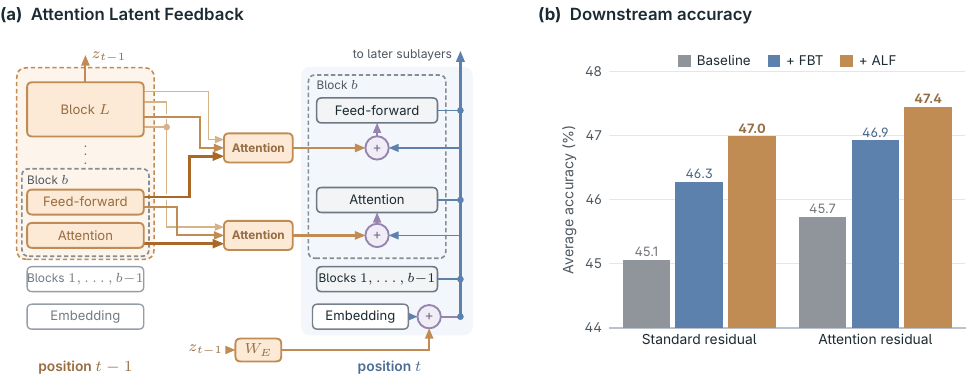}
  \caption[Attention Latent Feedback and downstream accuracy.]{\textbf{Attention Latent Feedback and downstream accuracy.}
  \textbf{(a)}~Each sublayer uses attention (orange) to weight previous-position outputs from its own depth upward
  and mix them with the current stream, while the projected top state enters the embedding sum.
  \textbf{(b)}~Domain-balanced accuracy on seventeen tasks for MoE-4.0B on a zoomed axis, comparing ALF with the
  Full-bandwidth Transformer (FBT); Tables~\ref{tab:downstream} and~\ref{tab:downstream_attnres} report full results.}
  \label{fig:teaser}
\end{figure}
}

\section{Introduction}
\label{sec:intro}

Autoregressive transformers \citep{vaswani2017attention} generate text one token at a time by passing each token through
a stack of residual blocks \citep{he2016deep} that progressively refines its representation from raw embeddings to
high-level context. Because decoding proceeds sequentially across tokens, every layer at the preceding position has
already finished before the first layer runs at the next. Standard causal self-attention, however, connects positions
strictly within the same layer, so shallow layers receive only the next token's embedding and have no access to the
preceding position's deeper hidden representations \citep{fan2020addressing}.

Recent architectural work relaxes this rigid information flow along two axes. Within a single position, depth-routing
designs replace the uniform residual sum by letting each layer weight earlier outputs individually, as in DenseFormer,
MUDDFormer, and Attention Residuals (AttnRes) \citep{pagliardini2024denseformer,xiao2025muddformer,kimiteam2026attention}.
Across positions, latent-feedback methods pass continuous states from one token to the next to improve loss and data
efficiency \citep{huang2026latentfeedback}: the Full-bandwidth Transformer (FBT) gates the preceding final state
into the next input, whereas Recycler adds a learned projection of it to the token embedding
\citep{wang2026full,bytedance2026recycler}.

\begin{wrapfigure}{r}{0.48\linewidth}
  \centering
  \vspace{-10pt}
  \includegraphics[width=\linewidth]{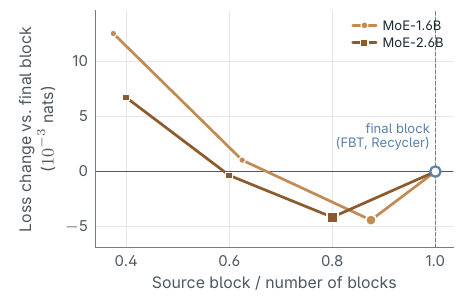}
  \caption[The final block is not the best feedback source.]{\textbf{The final block is not the best feedback source.}
  Held-out loss at the third evaluation pass is plotted relative to reading the final block, with one block's state
  projected and added to the next token input. Both models do best at a deep intermediate block,
  7 of 8 and 8 of 10,
  lowering loss by 4.4 and 4.1 $\times10^{-3}$ nats. The backbone, parameter
  count, FLOPs, checkpoint, and data order stay fixed, and each point is one run.}
  \label{fig:source_depth}
\end{wrapfigure}

An open question is whether the final layer is the best source for the state that is fed back. The final state is
trained to predict the next token, and the next position already receives that token as an embedding. It may therefore
be specialized for the output head and expose less intermediate computation. To test this, we swept the source block
of an additive input feedback path in MoE-1.6B and MoE-2.6B, with backbone, parameter count, FLOPs, checkpoint, and data
order fixed (Figure~\ref{fig:source_depth}). In both models the final block is not the best source. A deep intermediate
block lowers held-out loss by 4.4 and 4.1 $\times10^{-3}$ nats relative to the
final block, while the shallowest sources are clearly worse. Intermediate layers therefore carry information that
final-state feedback leaves unused. FBT replaces the addition with a gate, but it still reads only the final state.

Even with a better single source, feeding back only one state leaves two structural limitations. First, collapsing the
preceding position into one vector prevents weighting individual depths separately and obscures intermediate
representations with strong predictive signal. Second, injecting feedback only at the embedding denies later sublayers
direct access to deeper previous-position outputs. Other designs fix cross-position links in advance
\citep{huang2026latent,cai2026t2mlr,hao2026p2n,tang2026turbo}, while the Feedback Transformer \citep{fan2020addressing}
requires sequential training (Table~\ref{tab:survey}).

We introduce Attention Latent Feedback (ALF), which removes the need to choose a single source for the sublayer path.
Each sublayer keeps its current input while reading the preceding token's outputs at and above its depth
(Figure~\ref{fig:teaser}a), weighting them by content and mixing them at a learned share. This exposes every sublayer
to the intermediate computation that a single final state hides. To pair these reads with an input summary, ALF adds a
learned projection of the preceding top-layer state to the embedding, following Recycler. Both paths run in the final
quarter of pretraining via two parallel passes, while decoding stays single-pass.

We evaluate ALF on mixture-of-experts transformers at three scales under the Marin recipe \citep{marin2026}. On
MoE-1.6B, ALF lowers held-out loss by 0.024 nats over the standard residual, with similar gains at larger
scales, and outperforms FBT at all three scales under the same schedule. On MoE-4.0B, downstream accuracy rises from
45.1\% to 47.0\% (Figure~\ref{fig:teaser}b), and on AttnRes ALF delivers nearly the same loss
reduction across all three scales.

\noindent Our contribution is three-fold:
\begin{enumerate}[label=(\roman*),leftmargin=2.0em,itemsep=2pt,topsep=3pt,beginpenalty=10000,
  before={\interlinepenalty=10000}]
  \item We propose Attention Latent Feedback (ALF), which maps each previous-position sublayer output once through a
  low-rank projection and lets every sublayer attend over eligible depths via a learned softmax while pairing these
  reads with a projected top-layer input injection (\Cref{sec:method}).
  \item We show that ALF consistently lowers held-out loss across three MoE scales (MoE-1.6B to MoE-4.0B) on both
  standard and attention residual backbones, outperforms FBT at every scale, lifts MoE-4.0B downstream accuracy from
  45.1\% to 47.0\% (with matching gains under single-pass greedy generation across all 13
  completion tasks), and matches a 1.38$\times$ larger baseline token budget on MoE-1.6B
  (\Cref{sec:main,sec:attnres,sec:compute}).
  \item We further show that intermediate and nearby deeper outputs drive most of ALF's gain on both backbones, and
  that a fixed window over the current and next three sublayers retains 89\%--94\% of
  ALF's loss reduction at linear read cost (\Cref{sec:attnres,sec:maps,sec:window}).
\end{enumerate}

\begin{table}[t]
  \centering
  \newcommand{\hy}{\tikz[baseline=-.62ex]\fill[pageink!82] (0,0) circle (1mm);}
  \newcommand{\hp}{\tikz[baseline=-.62ex]{\draw[pageink!82, line width=.35pt] (0,0) circle (.95mm);
    \fill[pageink!82] (0,-.95mm) arc (-90:90:.95mm) -- cycle;}}
  \newcommand{\hn}{\tikz[baseline=-.62ex]\draw[pageink!55, line width=.35pt] (0,0) circle (.95mm);}
  \newcommand{\hna}{\textcolor{pagemuted}{--}}
  \newcommand{\two}[2]{\begin{tabular}[c]{@{}l@{}}#1\\[-3pt]{\scriptsize\color{pagemuted}#2}\end{tabular}}
  \newcommand{\fm}[1]{{\scriptsize\renewcommand{\arraystretch}{.92}\begin{tabular}[c]{@{}l@{}}#1\end{tabular}}}
  \newcommand{\hd}[1]{\begin{tabular}[b]{@{}c@{}}#1\end{tabular}}
  \captionsetup{position=top}
  \caption[Designs that change what a layer reads.]{\textbf{Designs that change what a layer reads.} Source gives the position and depths available to layer $l$ at $t$. Target identifies the receiver. Fixed weights are prescribed, static weights are learned and shared across tokens, and dynamic weights
  depend on the input. A single source requires no weighting across depths. Here $v_{t,i}$ is sublayer $i$'s
  output, $v_{t,0}=e_t$ before input fusion, and $x_{t,l}$ is the backbone input. The architecture specifies $s$
  and $S_l$. \hy~yes, \hp~in part, \hn~no, --~does not apply.}
  \label{tab:survey}
  \footnotesize
  \setlength{\tabcolsep}{2.0pt}
  \begin{tabular}{llccccc}
    \toprule
    Design & Input at position $t$ & Source & Target & \hd{Depth\\weights}
      & \hd{Parallel\\training} & \hd{One-pass\\decoding} \\
    \midrule
    \multicolumn{7}{l}{\textit{Residuals across depth}} \\
    \two{Standard residual}{\citet{he2016deep}}
      & \fm{$x_{t,l}=\sum_{i<l} v_{t,i}$} & $t,\ i<l$ & all & fixed & \hy & \hy \\[4pt]
    \two{DenseFormer}{\citet{pagliardini2024denseformer}}
      & \fm{$\sum_{i<l} w_{i\to l}\,v_{t,i}$} & $t,\ i<l$ & all & static & \hy & \hy \\[4pt]
    \two{Hyper-Connections, mHC}{\citet{zhu2024hyper}; \citet{xie2025mhc}}
      & \fm{$\sum_{k\le n} w_{t,k\to l}\,h^{k}_{t,l}$} & $t$, streams & all & \hna & \hy & \hy \\[4pt]
    \two{MUDDFormer}{\citet{xiao2025muddformer}}
      & \fm{$\sum_{i<l} w_{t,i\to l}\,v_{t,i}$} & $t,\ i<l$ & all & dynamic & \hy & \hy \\[4pt]
    \two{Attention Residuals, MSAR}{\citet{kimiteam2026attention}; \citet{qin2026multi}}
      & \fm{$\sum_{i<l} \alpha_{t,i\to l}\,v_{t,i}$} & $t,\ i<l$ & all & dynamic & \hy & \hy \\[4pt]
    \midrule
    \multicolumn{7}{l}{\textit{Feedback across positions}} \\
    \two{Feedback Transformer}{\citet{fan2020addressing}}
      & \fm{$x_{t,l}$ with attention to\\$\sum_i w_i\,v_{p,i}$} & $p<t$, all & all & static & \hn & \hy \\[4pt]
    \two{Full-bandwidth Transformer, Recycler}{\citet{wang2026full}; \citet{bytedance2026recycler}}
      & \fm{$v_{t,0}=g(e_t,v_{t-1,\mathrm{top}})$} & $t{-}1$, top & input & single & \hy & \hy \\[4pt]
    \two{PonderLM-2}{\citet{zeng2025ponderlm2}}
      & \fm{extra step, $v_{t,0}=v_{t-1,\mathrm{top}}$} & $t{-}1$, top & input & single & \hy & \hn \\[4pt]
    \two{Latent Recurrent Transformer}{\citet{huang2026latent}}
      & \fm{$x_{t,l}$ with memory $f_l(v_{t-1,s})$} & $t{-}1,\ s$ & all & single & \hp & \hy \\[4pt]
    \two{T$^2$MLR}{\citet{cai2026t2mlr}}
      & \fm{$g(x_{t,l},\,v_{t-1,s})$ at one $l<s$} & $t{-}1,\ s$ & one & single & \hy & \hy \\[4pt]
    \two{P2N}{\citet{hao2026p2n}}
      & \fm{$x_{t,l}+v_{t-1,s}$ at one $l<s$} & $t{-}1,\ s$ & one & single & \hy & \hy \\[4pt]
    \two{Turbo Connection}{\citet{tang2026turbo}}
      & \fm{$x_{t,l}+\sum_{s\in S_l} W_{s\to l}\,v_{t-1,s}$} & $t{-}1,\ S_l$ & several & \begin{tabular}[c]{@{}c@{}}static\\maps\end{tabular} & \hn & \hy \\[4pt]
    \two{Recirculation}{\citet{mozer2026recirculation}, at inference time}
      & \fm{$g(x_{t,l},\,v_{t-1,s})$ at one $l<s$} & $t{-}1,\ s$ & one & single & \hna & \hy \\[4pt]
    \midrule
    \rowcolor{laOrangeLine!9}
    \rule[-2.5ex]{0pt}{6.4ex}\two{\textbf{Attention Latent Feedback (ours)}$^{*}$}{Section~\ref{sec:method}}
      & \fm{$(1-\rho_l)\,x_{t,l}$\\${}+\rho_l\sum_{i\ge l} \beta_{t,i\to l}\,M_i\,v_{t-1,i}$}
      & $t{-}1,\ i\ge l$ & all & dynamic & \hy & \hy \\
    \bottomrule
  \end{tabular}\par
  \vspace{2pt}
  \begin{minipage}{\linewidth}\footnotesize
  $^{*}$ALF omits the stream-scale factor $c_{t,l}$ here for brevity and also adds a learned projection of the preceding
  top-layer state $z_{t-1}$ to the token embedding, $v_{t,0}=e_t+W_E\RMSNorm(z_{t-1})$, following Recycler.
  \end{minipage}
\end{table}

\section{Related work}
\label{sec:related}

\paragraph{Residual connections across depth.}
In a standard pre-norm transformer, the residual input to each layer is the sum of the token embedding and the
outputs below it \citep{he2016deep}. The first group in Table~\ref{tab:survey} shows how later designs change
that sum. DenseFormer learns weights for the shallower outputs \citep{pagliardini2024denseformer}, and
MUDDFormer lets those weights depend on the input \citep{xiao2025muddformer}. Hyper-Connections mix several
residual streams \citep{zhu2024hyper}, and manifold-constrained Hyper-Connections (mHC) restrict how those
streams mix \citep{xie2025mhc}. Hierarchical Residuals leave the summed outputs unchanged and widen the residual
state that carries them between blocks \citep{wang2026hierarchical}. Attention Residuals (AttnRes) use a learned query to weight the shallower
outputs \citep{kimiteam2026attention}. Multi-Stream Attention Residuals extend this depth weighting to
multiple streams \citep{qin2026multi}. In our experiments with Attention Latent Feedback (ALF), AttnRes weights
layers at the current position while ALF reads deeper outputs from the previous one.

\paragraph{Feedback across positions.}
The Feedback Transformer lets each layer attend to a memory that aggregates all depths from earlier positions, which
requires processing the sequence sequentially during training \citep{fan2020addressing}. Later designs draw feedback
from the immediately preceding position, with the source and destination fixed by the architecture. The Full-bandwidth
Transformer (FBT) gates the final state into the next token input \citep{wang2026full}, Recycler adds a projection of
it to the embedding \citep{bytedance2026recycler}, and PonderLM-2 uses it for an extra latent step
\citep{zeng2025ponderlm2}. \citet{huang2026latentfeedback} study the data efficiency of additive feedback when the same
training examples appear repeatedly. Within the stack, the Latent Recurrent Transformer gives each block a memory drawn
from one chosen depth \citep{huang2026latent}. T$^2$MLR feeds a middle layer into an earlier one \citep{cai2026t2mlr},
while P2N links the end of a chosen block range to its start at the next position \citep{hao2026p2n}. Turbo Connection
uses several prescribed links \citep{tang2026turbo}, and Recirculation introduces a link at inference time
\citep{mozer2026recirculation}. While ALF retains Recycler's additive input connection, inside the stack each sublayer
dynamically weights every output at or above its own depth from the preceding position.

\paragraph{Parallel training across positions.}
Parallelizing recurrent models over sequence length began with parallel prefix scans for linear recurrences
\citep{martin2018parallelizing} and extended to nonlinear sequential models through fixed-point solvers such as Jacobi
iteration \citep{song2021accelerating}, parallel Newton iteration in DEER \citep{lim2024parallelizing}, and stabilized
quasi-Newton updates in ELK \citep{gonzalez2024towards}. In feedback transformers, FBT and Recycler use successive
Jacobi passes to feed back the final state \citep{wang2026full,bytedance2026recycler}, while P2N restricts its Jacobi
updates to a middle group of layers \citep{hao2026p2n}. Each pass carries feedback one token farther along the sequence.
ALF trains with two parallel Jacobi passes, whereas during decoding the preceding token's states are already cached and
each new token requires only one pass.

\paragraph{Looped transformers.}
Looped transformers apply a shared block repeatedly at the same position \citep{dehghani2019universal}. More
iterations give a token more computation \citep{geiping2025scaling}, and a learned exit gate can decide when to stop
\citep{zhu2025ouro}. \citet{schwethelm2026recurrence} compare repeated blocks with distinct layers at fixed unrolled
depth, and at matched training compute full backpropagation through fewer loops can outperform truncation through more
loops \citep{xiong2026truncate}. Inkling 2 Mini \citep{thinkingmachines2026inkling2mini} likewise loops its middle
blocks across three passes per token using block-wise attention residuals. To avoid sequential loop latency during
decoding, the Parallel Loop Transformer computes different loops for different tokens in parallel within one pass while
sharing key-value caches across loops \citep{wu2025plt}, and a shallow loop can draft a token using full-depth values
from the verified prefix \citep{li2026das}. Whereas looped transformers reuse weights to
add compute at the same position during decoding, ALF reuses deeper representations already computed at the preceding
position within a single decoding pass.

\termtinttrue
\section{Method}
\label{sec:method}

We consider a pre-norm transformer \citep{he2016deep,vaswani2017attention} with $L$ blocks and number its
$2L$ sublayers from $1$ at the bottom, so that block $b\in\{1,\dots,L\}$ holds attention sublayer $2b-1$
and feed-forward sublayer $2b$, while index $0$ denotes the token input $v_{t,0}$ (with $x_{t,1}=v_{t,0}$).
We call $v_{t,0},\dots,v_{t,2L}$ the sources at position $t$ and $i$ the depth of source $v_{t,i}$.
Throughout, $l=1,\dots,2L$ indexes the receiving sublayer, $x_{t,l}$ is its backbone input, and
$z_t$ denotes the final state before output normalization ($x_{t,2L+1}$ on standard residuals, or the pre-head depth
mixture on AttnRes).

\subsection{Overview}
\label{sec:overview}

\begin{figure}[t]
  \centering
  \includegraphics{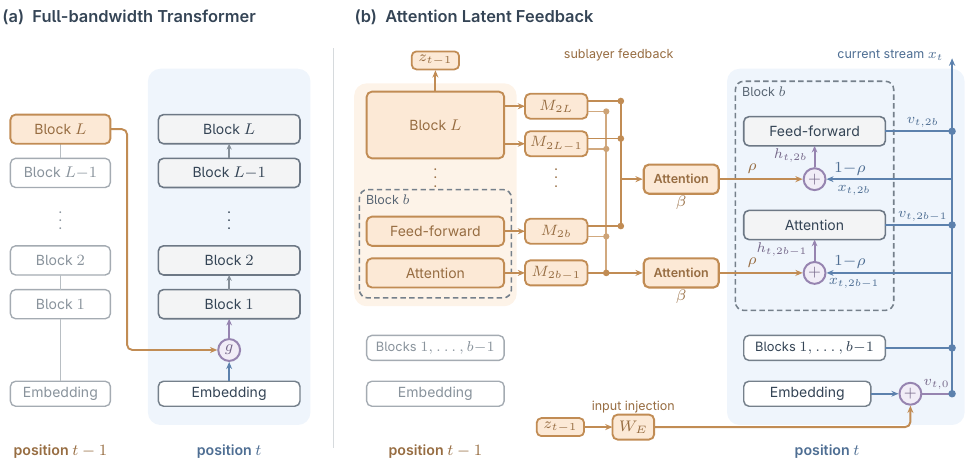}
  \caption{\textbf{The Full-bandwidth Transformer and Attention Latent Feedback.} Blue marks the current position,
  orange what is read from the previous position, and purple the mixtures where they meet. \textbf{(a)}~FBT fuses block
  $L$ of position $t-1$ into the token embedding of position $t$ through a learned gate $g$. \textbf{(b)}~With ALF,
  each sublayer of block $b$ attends over the mapped outputs of blocks $b$ to $L$ (weights $\beta$) and mixes the result
  into its backbone input at share $\rho$. The final state $z_{t-1}$ also joins the token embedding through $W_E$
  \eqref{eq:embed-inject}. Faded blocks are not read.}
  \label{fig:arch}
\end{figure}

\paragraph{Reusing deeper states across positions.}
In an autoregressive transformer, every block has finished at position $t-1$ before decoding starts at
position $t$. A shallow attention sublayer at the new position still cannot read those deeper outputs,
because standard self-attention restricts it to keys and values formed at its own depth. Attention Latent
Feedback (ALF) lets every attention and feed-forward sublayer reuse that computation. Each sublayer keeps
its usual input from the sublayers below it and also reads its own depth and every deeper one from position
$t-1$. Figure~\ref{fig:arch} compares ALF with the Full-bandwidth Transformer (FBT), which feeds only the
final state back to the token input.

\paragraph{Mixing the two streams.}
Sublayer $l$ draws on two streams. The \cstream{} $x_{t,l}$ is the input that the backbone assembles
without ALF from the sources $i<l$ at position $t$, whereas the \fbstream{} carries the sources $i\ge l$ from
position $t-1$ (Figure~\ref{fig:arch}b). When sublayer $l$ runs, position $t$ has not yet computed depth
$l$ or any depth above it, so the two streams together cover every depth in its most recent version. In
training, an earlier pass supplies the outputs at $t-1$ (\Cref{sec:training}). Combining the two streams yields
the mixed sublayer input
\begingroup
\addtolength{\abovedisplayskip}{14pt}%
\addtolength{\abovedisplayshortskip}{14pt}%
\begin{equation}
  h_{t,l}
  =(1-\rho_l)\,\eqcallout[laBlueLine][0pt][eqx1]{current stream}{x_{t,l}}
  +\rho_l\,c_{t,l}\sum_{i=l}^{2L}\wbeta_{t,i\to l}\,M_i\,\eqcallout[laOrangeLine][0pt][eqv1]{sublayer feedback}{v_{t-1,i}}.
  \label{eq:mix}
\end{equation}
\endgroup
Here $M_i$ is a learned low-rank linear map of source $v_{t-1,i}$ and $\wbeta_{t,i\to l}$ is a softmax
weight over depth, both detailed in \Cref{sec:feedback}, while $\rho_l=\sigma(a_l)\in(0,1)$ is a learned
per-sublayer mixing share parameterized through the logistic sigmoid of a scalar $a_l$. The rescaling scalar
$c_{t,l}$ matches the feedback sum's root mean square (RMS) to $x_{t,l}$ on standard residuals and equals
$1$ on Attention Residuals (AttnRes), as detailed in \eqref{eq:scale}. The \cstream{} aggregates the shallower sources at position $t$ according to the backbone,
\begin{equation}
  \eqbox[laBlueLine]{eqx2}{x_{t,l}}=
  \begin{cases}
    \sum_{i=0}^{l-1}v_{t,i}, & \text{Standard residual},\\[4pt]
    \sum_{i=0}^{l-1}\walpha_{t,i\to l}\,v_{t,i}, & \text{Attention residual},
  \end{cases}
  \label{eq:xupdate}
\end{equation}
where standard residuals take an unweighted sum and AttnRes takes a depth-softmax average whose weights
$\walpha_{t,i\to l}$ are defined in \Cref{app:model}. Each sublayer then applies pre-normalization to the
mixed input and produces its output,
\begin{equation}
  \eqbox[laOrangeLine]{eqv2}{v_{t,l}}=
  \begin{cases}
    \operatorname{Attn}_b\bigl(\RMSNorm(h_{1:t,l})\bigr), & l=2b-1,\\[4pt]
    \operatorname{FFN}_b\bigl(\RMSNorm(h_{t,l})\bigr), & l=2b,
  \end{cases}
  \label{eq:vupdate}
\end{equation}
where $\RMSNorm$ is applied positionwise and causal self-attention attends over
$h_{1:t,l}=(h_{1,l},\dots,h_{t,l})$, whereas the feed-forward sublayer acts on $h_{t,l}$ alone. Because only
$v_{t,l}$ enters deeper current-stream inputs $x_{t,l'}$ ($l'>l$), $h_{t,l}$ stays local to sublayer $l$,
while $v_{t,l}$ enters sublayer feedback at position $t+1$ for depth $l$ and below.

\paragraph{Token-input injection.}
Even with sublayer feedback in place, both backbones also inject the previous position's final state into the
token input. \emph{Whereas sublayer feedback exposes individual sublayer outputs $v_{t-1,i}$, the final latent state
$z_{t-1}$ carries complementary information: it is the accumulated representation that directly produces the
next-token prediction logits and probabilities at position $t-1$, preserving richer distributional context
than the single sampled token alone.} Let $e_t$ denote the token embedding at position $t$ and
$W_E\in\mathbb{R}^{d\times d}$ a learned projection. Following Recycler \citep{bytedance2026recycler} and
\citet{huang2026latentfeedback}, we initialize the current stream and depth-0 source as
\begin{equation}
  x_{t,1}=v_{t,0}=e_t+W_E\,\RMSNorm(z_{t-1}).
  \label{eq:embed-inject}
\end{equation}
Here $\RMSNorm$ is parameter-free (no learned gain), and the last feed-forward output $v_{t-1,2L}$ remains
a separate source for the sublayer feedback because $z_{t-1}$ represents the accumulated residual state
rather than that single sublayer's output. We initialize $W_E$ to zero so that training starts from the
unmodified embedding, and scale its learning rate inversely with model width $d$ to maintain scale-invariant
updates across model sizes (\Cref{app:injection}).

\paragraph{Why feedback starts at depth $l$.}
Restricting the sum in \eqref{eq:mix} to $i\ge l$ roughly halves the number of sources each sublayer scores
compared to attending over all $2L$ depths. It also omits shallower outputs $i<l$ from position $t-1$,
whose information is already reachable through causal self-attention in the sublayers below $l$. What makes
cross-position feedback valuable is access to deeper computation from the preceding token, which standard
attention never exposes to shallower sublayers \citep{fan2020addressing,hao2026p2n,tang2026turbo}.

\subsection{The sublayer feedback}
\label{sec:feedback}

\paragraph{Low-rank source maps.}
Each source output $v_{t-1,i}$ was originally produced for position $t-1$, so a sublayer at position $t$
benefits from adapting it before reading. Motivated by LoRA \citep{hu2022lora}, we apply a residual
low-rank map $M_i=I+U_iD_i$ to each source output once and reuse the mapped vector at every sublayer that
reads it, where $d$ is the model width, $U_i\in\mathbb{R}^{d\times r}$ and $D_i\in\mathbb{R}^{r\times d}$.
The identity term $I$ preserves all information in the original output, while the low-rank projection
$U_iD_i$ adapts a compact subspace on top of it without paying for a full $d\times d$ matrix. We set
$r=64$ so that the maps add less than one percent to a forward pass.

\paragraph{Tied key-value attention over depth.}
Each receiving sublayer $l$ selects among its eligible sources using a learned query $u_l\in\mathbb{R}^d$.
Following Attention Residuals \citep{kimiteam2026attention}, we tie the keys and values by reusing the
same mapped output $M_iv_{t-1,i}$ for both scoring and aggregation instead of adding separate key and
value projections. Tying them avoids extra projection parameters at every depth and lets the query score
the exact representation that will be mixed into the sublayer, while normalizing the key prevents
large-magnitude sources from dominating the scores. A softmax over eligible source depths gives the
weights
\begin{equation}
  \wbeta_{t,i\to l}=\frac{\exp\bigl(u_l^{\top}\RMSNorm(M_iv_{t-1,i})/\sqrt{d}\bigr)}
                        {\sum_{i'=l}^{2L}\exp\bigl(u_l^{\top}\RMSNorm(M_{i'}v_{t-1,i'})/\sqrt{d}\bigr)}.
  \label{eq:hsoftmax}
\end{equation}
Although each sublayer shares one query across positions, the resulting weights adapt dynamically to the
source outputs. The weighted sum $n_{t,l}=\sum_{i=l}^{2L}\wbeta_{t,i\to l}M_iv_{t-1,i}$ combines the
unnormalized mapped outputs. We initialize $U_i$ and $u_l$ to zero, so each map starts at the identity and
each sublayer weights its eligible sources uniformly, and set $a_l$ so that the initial share is
$\rho_l=0.01$.

\paragraph{Matching stream scales across depth.}
Because its softmax weights sum to one, the feedback mixture $n_{t,l}$ stays roughly constant in scale across
depth, whereas on a standard residual backbone the unweighted sum $x_{t,l}$ accumulates across layers and
grows in RMS with depth. Mixing them directly would therefore attenuate the feedback in deeper sublayers and
give $\rho_l$ a different effective strength at each depth. We match the two streams by setting
\begin{equation}
  c_{t,l}=
  \begin{cases}
    \dfrac{\operatorname{RMS}(x_{t,l})}{\operatorname{RMS}(n_{t,l}) + \epsilon}, & \text{Standard residual},\\[8pt]
    1, & \text{Attention residual},
  \end{cases}
  \label{eq:scale}
\end{equation}
where $\operatorname{RMS}(x)=\|x\|_2/\sqrt{d}$ is the root mean square of $x\in\mathbb{R}^d$ and $\epsilon>0$ is
a small constant for numerical stability. On standard residuals, this rescaling brings both terms to the same
RMS before mixing so that $\rho_l$ controls their balance uniformly across depth. On AttnRes, $x_{t,l}$ is
itself a depth-softmax average (\Cref{app:model}), so both streams already share the same scale and need no
rescaling.

\subsection{Training and inference}
\label{sec:training}
\label{sec:inference}

\begin{figure}[t]
  \centering
  \includegraphics{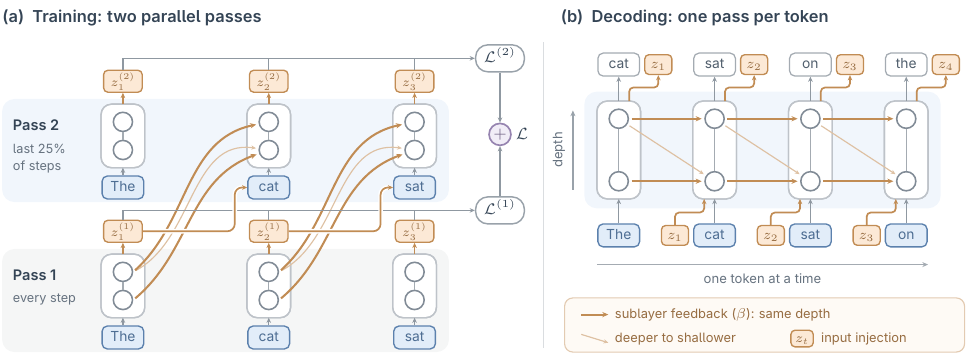}
  \caption{\textbf{Parallel training and decoding.} Each column is one forward pass of the backbone at one token, its
  circles are the sublayers with depth increasing upward, the blue chips are the input tokens, and the orange chips are
  the final states that the next position adds to its token input \eqref{eq:embed-inject}. \textbf{(a)}~Training runs
  two passes over all positions. Pass~1 is the plain backbone, while in pass~2 every sublayer reads the outputs that
  pass~1 left at the preceding position, at its own depth (solid) and at every deeper sublayer (light). Each pass
  applies a next-token loss to its final states across all positions, and the two are added \eqref{eq:loss}. \textbf{(b)}~Decoding runs one pass per token and reads the preceding token's cached states.}
  \label{fig:passes}
\end{figure}

\paragraph{Two-pass Jacobi training.}
Because both feedback paths at position $t$ read the outputs of position $t-1$, exact evaluation is a nonlinear
recurrence across positions. To parallelize training over the sequence length, we adopt Jacobi fixed-point iteration
\citep{song2021accelerating}, part of the broader family of parallel fixed-point solvers for nonlinear sequential models
\citep{lim2024parallelizing,gonzalez2024towards} and also used in recent feedback transformers
\citep{wang2026full,bytedance2026recycler}. Given a token sequence $(y_1,\dots,y_{T+1})$, pass~1 runs the plain
backbone across all $T$ positions with both feedback paths disabled, setting $x^{(1)}_{t,1}=v^{(1)}_{t,0}=e_t$ and
$h^{(1)}_{t,l}=x^{(1)}_{t,l}$. On each subsequent pass $k=2,\dots,K$, the \cstream{} $x^{(k)}_{t,l}$ is computed within
pass~$k$ by \eqref{eq:xupdate}, while the input injection \eqref{eq:embed-inject} and sublayer mixing \eqref{eq:mix}
read the shifted outputs from pass~$k-1$ across all positions in parallel (Figure~\ref{fig:passes}a):
\begin{equation}
  \begin{aligned}
    x^{(k)}_{t,1} &= v^{(k)}_{t,0} = e_t + W_E\,\RMSNorm\bigl(z^{(k-1)}_{t-1}\bigr), \\
    h^{(k)}_{t,l} &= (1-\rho_l)\,x^{(k)}_{t,l}
      + \rho_l\,c^{(k)}_{t,l}\sum_{i=l}^{2L}\wbeta^{(k)}_{t,i\to l}\,M_i\,v^{(k-1)}_{t-1,i},
  \end{aligned}
  \label{eq:jacobi}
\end{equation}
where $\wbeta^{(k)}_{t,i\to l}$ and $c^{(k)}_{t,l}$ follow \eqref{eq:hsoftmax} and
\eqref{eq:scale} using $x^{(k)}_{t,l}$ and $v^{(k-1)}_{t-1,i}$, and each sublayer produces $v^{(k)}_{t,l}$ via
\eqref{eq:vupdate}. At the start of any document (including $t=1$), both feedback terms are disabled so that feedback
never crosses document boundaries. By induction on $k$, pass~$k$ matches exact sequential evaluation on the first $k$
tokens of every document, advancing exact feedback by one token per pass.

Let $p^{(k)}_t$ denote the next-token distribution predicted from the final state $z^{(k)}_t$ at position $t$ on
pass~$k$. Each of the $K\ge2$ passes incurs a sequence-averaged next-token cross-entropy $\mathcal{L}^{(k)}$, and we
optimize their sum with equal weight on the first pass and the mean of the feedback passes:
\begin{equation}
  \mathcal{L}=\mathcal{L}^{(1)}+\frac{1}{K-1}\sum_{k=2}^{K}\mathcal{L}^{(k)},
  \qquad
  \mathcal{L}^{(k)}=-\frac{1}{T}\sum_{t=1}^{T}\log p^{(k)}_t(y_{t+1}).
  \label{eq:loss}
\end{equation}
We backpropagate through all passes of \eqref{eq:jacobi} without detaching $v^{(k-1)}_{t-1,i}$ or $z^{(k-1)}_{t-1}$, so
the feedback-pass losses train earlier passes to produce useful feedback representations while $\mathcal{L}^{(1)}$ keeps
the plain backbone well-calibrated. In our main experiments, we train the backbone with a single pass for the first
$75\%$ of steps and then enable feedback with $K=2$ passes for the final quarter, which is the schedule used in
\Cref{sec:results} and \Cref{sec:compute}. Two passes already suffice in practice: every sublayer receives cross-position
feedback from pass~1, adding a third training pass yields negligible improvement over $K=2$ at higher training compute,
and evaluating further parallel passes stabilizes held-out loss rapidly (\Cref{sec:passes} and
\Cref{app:passes}).

\paragraph{Single-pass decoding and prompt prefill.}
During autoregressive decoding, ALF caches the $2L$ sublayer outputs and the top-layer state of the preceding token, so
a single forward pass at each new token evaluates both feedback paths exactly (Figure~\ref{fig:passes}b). At the start
of a document there is no preceding token, and both paths are disabled. For a known sequence, such as a prompt during prefill or a document during teacher-forced evaluation, multiple parallel
Jacobi passes can propagate feedback across all positions, or a prompt can be prefilled with a single plain backbone
pass before enabling feedback on the continuation. We examine how loss evolves across passes in \Cref{sec:passes} and compare both prefill strategies in
\Cref{sec:recipe}.

\termtintfalse

\newpage
\begin{table}[t]
  \noindent
  \begin{minipage}[b]{.485\textwidth}
    \centering
    \footnotesize
    \setlength{\tabcolsep}{4pt}
    \begin{tabular}{l rrr}
\toprule
Model & MoE-1.6B & MoE-2.6B & MoE-4.0B \\
\midrule
Standard residual & 2.0577 & 1.9400 & 1.8456 \\
\quad + FBT & 2.0447 & 1.9261 & 1.8322 \\
\rowcolor{laOrangeLine!25}\quad + ALF & \textbf{2.0333} & \textbf{1.9170} & \textbf{1.8238} \\
\midrule
Attention residual & 2.0424 & 1.9244 & 1.8312 \\
\quad + FBT & 2.0336 & 1.9144 & 1.8227 \\
\rowcolor{laOrangeLine!25}\quad + ALF & \textbf{2.0193} & \textbf{1.9020} & \textbf{1.8097} \\
\bottomrule
\end{tabular}
\par\vspace{0pt}
  \end{minipage}\hfill
  \begin{minipage}[b]{.485\textwidth}
    \centering
    \includegraphics[width=\linewidth]{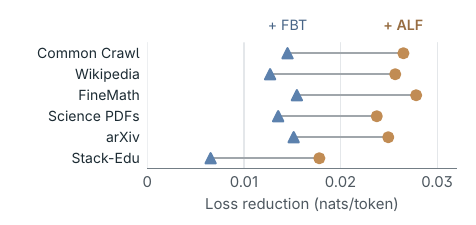}\par\vspace{0pt}
  \end{minipage}
  \par\noindent
  \begin{minipage}[t]{.485\textwidth}
    \vspace{0pt}
    \captionsetup{position=bottom,width=\linewidth}
    \caption[Held-out loss on two backbones.]{\textbf{Held-out loss on two backbones.} Loss in nats per token,
    averaged over six Dolma 3 domains. ALF denotes Attention Latent Feedback. FBT uses its stronger recipe without
    prefix mixin (\Cref{sec:recipe}, \Cref{app:setup}), and both feedback methods use three evaluation passes.}
    \label{tab:main}
  \end{minipage}\hfill
  \begin{minipage}[t]{.485\textwidth}
    \vspace{0pt}
    \captionsetup{type=figure,position=bottom,width=\linewidth}
    \caption[Loss reductions across six domains.]{\textbf{Loss reductions across six domains.}
    MoE-1.6B runs from Table~\ref{tab:main} on the standard residual. Points show baseline minus method loss
    in nats per token. Blue triangles denote FBT and orange circles denote ALF. Larger gains are better.}
    \label{fig:domains}
  \end{minipage}
\end{table}

\section{Experimental results}
\label{sec:results}

\subsection{Experimental setup}
\label{sec:setup}

\paragraph{Models and data.}
We pretrain a ladder of three mixture-of-experts transformers following the Marin recipe \citep{marin2026}, using
Marin's MoE-1.6B and MoE-4.0B configurations together with an interpolated MoE-2.6B model (Table~\ref{tab:models}).
Each model trains on 791 tokens per activated parameter from our cleaned Dolma 3 mixture
\citep{olmo2025olmo3}.

\begin{table}[ht]
  \centering
  \captionsetup{position=top}
  \caption[Model configurations and training budgets.]{\textbf{Model configurations and training budgets.} The three
  MoE models are named by their total parameters; MoE-2.6B interpolates between Marin's MoE-1.6B and MoE-4.0B
  configurations. Every block routes each token to 8 of 384 experts plus two shared
  experts, and activated parameters exclude embeddings. Batch counts sequences of 4{,}096 tokens.}
  \label{tab:models}
  \footnotesize
  \setlength{\tabcolsep}{5pt}
  \begin{tabular}{@{}l rr rrrr r@{}}
\toprule
 & \multicolumn{2}{c}{Parameters} & & & & & \\
\cmidrule(lr){2-3}
Model & Total & Activated & Blocks & Heads & Width & Batch & Tokens \\
\midrule
MoE-1.6B & 1.59B & 61M & 8 & 6 & 768 & 1{,}024 & 47.9B \\
MoE-2.6B & 2.60B & 102M & 10 & 7 & 896 & 1{,}440 & 80.7B \\
MoE-4.0B & 3.98B & 162M & 12 & 8 & 1{,}024 & 2{,}048 & 128.1B \\
\bottomrule
\end{tabular}

\end{table}

\paragraph{Training protocol.}
We train the standard-residual and AttnRes backbones from scratch, then branch FBT \citep{wang2026full} and ALF from
the $75\%$ checkpoint of each. In each paired comparison, the two branches see the same data order, run two passes
over the remaining quarter, and follow the same backbone learning-rate schedule (\Cref{app:setup}), while the continued
backbone runs a single pass over the same data. FBT retains the original paper's gated input fusion and omits the
prefix mixin, matching its stronger held-out loss recipe in \Cref{sec:recipe}, while ALF omits both the prefix mixin
and jitter (baseline details and recipe ablations appear in \Cref{app:setup} and \Cref{app:passes}). The main
comparison uses the standard residual, and \Cref{sec:attnres} extends it to AttnRes; training trajectories appear in
Figure~\ref{fig:training} (\Cref{app:setup}).

\begin{table}[t]
  \centering
  \captionsetup{position=top}
  \caption[Downstream tasks on MoE-4.0B.]{\textbf{Downstream tasks on MoE-4.0B.} Accuracy in percent (higher is better)
  and perplexity of the correct answer (lower is better) on the seventeen tasks of \Cref{sec:setup}, grouped by
  domain, with the best result of each row in bold. Each domain ends with its average, and the last row weights the
  four domains equally (\Cref{app:eval}). ALF leads the standard residual in all four domains, raising the average from
  45.1 to 47.0 points.}
  \label{tab:downstream}
  \footnotesize
  \setlength{\tabcolsep}{6pt}
  \begin{tabular}{@{}l ccc ccc@{}}
\toprule
 & \multicolumn{3}{c}{\textbf{Accuracy (\%) $\uparrow$}} & \multicolumn{3}{c}{\textbf{Perplexity $\downarrow$}} \\
\cmidrule(lr){2-4}\cmidrule(l){5-7}
Task & Standard & + FBT & + ALF & Standard & + FBT & + ALF \\
\midrule
\multicolumn{7}{@{}l}{\textit{Knowledge}} \\
TriviaQA & 20.2 & 20.5 & \textbf{20.9} & 12.36 & 12.52 & \textbf{12.18} \\
Natural Questions & \textbf{7.0} & \textbf{7.0} & 6.9 & 19.47 & \textbf{18.15} & 18.56 \\
WebQuestions & 12.8 & \textbf{13.9} & 13.5 & 16.15 & \textbf{15.56} & 15.65 \\
Jeopardy & 33.9 & 36.2 & \textbf{36.3} & 4.35 & 4.16 & \textbf{4.11} \\
QA-Wikidata & 63.2 & 62.9 & \textbf{63.9} & 4.04 & 4.00 & \textbf{3.93} \\
ARC-Easy & 72.2 & 72.9 & \textbf{73.6} & \textbf{7.92} & 7.99 & 8.00 \\
ARC-Challenge & 45.5 & \textbf{45.7} & 45.5 & 11.46 & 11.41 & \textbf{11.39} \\
\cmidrule(lr){1-7}
Average & 36.4 & 37.0 & \textbf{37.2} & 9.34 & 9.15 & \textbf{9.11} \\
\midrule
\multicolumn{7}{@{}l}{\textit{Reading comprehension}} \\
LAMBADA & 60.2 & 61.0 & \textbf{61.8} & 4.35 & 4.26 & \textbf{4.19} \\
SQuAD & 44.6 & 45.5 & \textbf{46.6} & 1.91 & 1.90 & \textbf{1.86} \\
SQuAD 2.0 & 37.5 & 37.4 & \textbf{40.1} & 2.14 & 2.15 & \textbf{2.07} \\
TyDi QA & 39.5 & 44.5 & \textbf{45.2} & 1.61 & 1.59 & \textbf{1.57} \\
CoQA & 32.6 & 33.8 & \textbf{34.4} & 3.53 & 3.45 & \textbf{3.35} \\
DROP & 14.4 & 15.1 & \textbf{16.1} & 6.90 & 6.76 & \textbf{6.69} \\
\cmidrule(lr){1-7}
Average & 38.1 & 39.6 & \textbf{40.7} & 2.98 & 2.94 & \textbf{2.88} \\
\midrule
\multicolumn{7}{@{}l}{\textit{Commonsense reasoning}} \\
HellaSwag & 59.8 & 60.6 & \textbf{61.2} & 11.96 & 11.77 & \textbf{11.69} \\
PIQA & 75.4 & 74.8 & \textbf{76.1} & 12.46 & 12.23 & \textbf{12.11} \\
\cmidrule(lr){1-7}
Average & 67.6 & 67.7 & \textbf{68.7} & 12.21 & 12.00 & \textbf{11.90} \\
\midrule
\multicolumn{7}{@{}l}{\textit{Mathematics}} \\
ASDiv & 36.4 & 39.4 & \textbf{40.3} & 4.93 & 4.52 & \textbf{4.43} \\
MAWPS & 40.0 & 42.3 & \textbf{42.4} & 4.05 & 3.81 & \textbf{3.77} \\
\cmidrule(lr){1-7}
Average & 38.2 & 40.8 & \textbf{41.3} & 4.46 & 4.15 & \textbf{4.08} \\
\midrule
\textbf{All domains} & 45.1 & 46.3 & \textbf{47.0} & 6.24 & 6.05 & \textbf{5.97} \\
\bottomrule
\end{tabular}

\end{table}

\paragraph{Evaluation protocol.}
Our primary held-out and downstream benchmarks (Tables~\ref{tab:main} and~\ref{tab:downstream}) use \emph{parallel
teacher-forced scoring}, where every token of a sequence is given up front and processed in parallel rather than one
token at a time. To avoid propagating feedback sequentially across tokens, FBT and ALF keep feedback enabled as in
\citet{wang2026full} and run three parallel passes over each sequence (\Cref{sec:training}), whereas the
continued backbone needs only a single pass. All loss, perplexity and accuracy metrics are computed from the final
pass's output distribution. For held-out language modeling, we evaluate next-token loss over whole documents at the
final checkpoint and average in nats per token across six Dolma 3 domains, with 0.003 nats as our single-scale
ablation threshold (\Cref{app:setup}). Because additional Jacobi passes further lower ALF's loss while raising FBT's
(\Cref{sec:passes}), three-pass scoring is conservative for ALF, and prompt prefill is compared in \Cref{sec:recipe}.
In \Cref{sec:main}, we complement these scores with true single-pass greedy autoregressive generation across all
13 completion tasks (Table~\ref{tab:generation}).

\paragraph{Downstream evaluation.}
We evaluate seventeen tasks spanning knowledge, reading comprehension, commonsense reasoning and mathematics under
the same three-pass parallel scoring, using the prompts and few-shot counts of \citet{schwethelm2026recurrence} and
DCLM CORE \citep{li2024datacomp} (\Cref{app:eval}). Following \citet{schwethelm2026recurrence}, each input concatenates
the few-shot examples, question and candidate answer, and only the answer tokens are scored on the final pass. From the
final pass, we compute both reference-answer perplexity, a finer-grained measure of base-model quality
\citep{heineman2025signal}, and accuracy: multiple-choice tasks select the option with the lowest mean negative
log-likelihood per scored token, whereas completion tasks require the top-1 next token to match the reference at every
answer position. Domain averages weight tasks equally, and the overall average weights domains equally. Task
selection, checkpoints and uncertainty criteria appear in \Cref{app:eval}.

\subsection{Main results}
\label{sec:main}

\paragraph{Held-out language modeling loss.}
Across all three model scales on standard residuals, ALF lowers held-out loss relative to both the continued backbone
and FBT under the same two-pass schedule (Table~\ref{tab:main}). On MoE-1.6B, it reduces the standard residual's loss by
0.024 nats on matched data, and the gain persists without attenuating on MoE-2.6B and MoE-4.0B. On MoE-1.6B,
this gain spans all six validation domains (from code and mathematics to scientific papers and web text), with ALF
ahead of FBT in every domain (Figure~\ref{fig:domains}). Although FBT also improves on the backbone, ALF remains at
least 0.008 nats lower at every scale, and we compare these gains against compute in \Cref{sec:compute}.

\paragraph{Downstream task performance.}
These held-out gains carry over directly to downstream benchmarks on MoE-4.0B (Table~\ref{tab:downstream}). ALF
raises the four-domain average accuracy from 45.1 to 47.0 points and improves accuracy and
reference-answer perplexity on nearly every individual task, with the largest gains in mathematics and reading
comprehension, where reusing deeper context across tokens is most valuable. Relative to FBT, ALF adds
0.7 points on average and leads in all four domains, with the largest margins in reading comprehension
and commonsense reasoning (\Cref{app:eval}).

\paragraph{Autoregressive generation.}
Because Table~\ref{tab:downstream} scores reference answers under parallel teacher forcing, we also evaluate true
greedy autoregressive decoding on MoE-4.0B across all 13 completion benchmarks from Table~\ref{tab:downstream}
using cached key--value and previous-token feedback states (Table~\ref{tab:generation}; protocol and per-task scores in
\Cref{app:eval}). Each generated token runs a single forward pass reading the completed previous token's states.
Under two-pass \emph{Fused} prompt prefill, ALF lifts the three-domain average EM from 37.45\% to 39.73\%
($+2.28$ points, with F1 rising from 43.08\% to 45.57\%) and outperforms the standard residual on
all 13/13 tasks. It also leads FBT (38.50\%) by 1.22 points, while the
linear-read fixed nearest-four variant (which averages the four nearest eligible sublayer outputs in place of a learned
depth softmax; \Cref{sec:window}) reaches 39.35\% ($+1.90$ points over the backbone), confirming that
per-sublayer depth-selective reading carries over to real generation.

\begin{table}[!ht]
  \centering
  \captionsetup{position=top}
  \caption[Autoregressive generation on MoE-4.0B.]{\textbf{Autoregressive generation on MoE-4.0B.}
  Greedy single-pass decoding accuracy (normalized EM and token F1, in percent, higher is better) across all
  13 completion benchmarks from Table~\ref{tab:downstream}, using cached KV and previous-token feedback
  states (\Cref{app:eval}). \emph{Fused prefill} runs two parallel passes over the prompt before single-pass
  autoregressive decoding; \emph{Soft prefill} runs one plain prompt pass (\Cref{sec:recipe}). The standard residual has
  no feedback path, so its Fused and Soft columns coincide. Domain averages weight the three domains equally (Knowledge,
  Reading Comprehension, Mathematics), and bold marks the highest score in each column.}
  \label{tab:generation}
  \small
  \setlength{\tabcolsep}{6.5pt}
  \begin{tabular}{l rrrrr rr}
\toprule
 & \multicolumn{5}{c}{\textbf{Fused prefill (2-pass)} $\uparrow$} & \multicolumn{2}{c}{\textbf{Soft prefill (1-pass)} $\uparrow$} \\
\cmidrule(lr){2-6}\cmidrule(l){7-8}
Model & Know. & RC & Math & Avg.\ EM & Avg.\ F1 & Avg.\ EM & Avg.\ F1 \\
\midrule
Standard residual & 32.37 & 40.59 & 39.38 & 37.45 & 43.08 & 37.45 & 43.08 \\
+ FBT & 32.46 & 42.10 & 40.94 & 38.50 & 44.29 & 37.48 & 43.23 \\
\rowcolor{laOrangeLine!12}+ Fixed nearest-4 & \textbf{33.12} & 43.06 & 41.87 & 39.35 & 45.09 & 37.54 & 43.24 \\
\rowcolor{laOrangeLine!25}+ ALF & 33.06 & \textbf{43.35} & \textbf{42.76} & \textbf{39.73} & \textbf{45.57} & \textbf{38.01} & \textbf{43.70} \\
\bottomrule
\end{tabular}

\end{table}

Under one-pass plain (\emph{Soft}) prompt prefill, ALF still achieves the highest EM and F1 (38.01\% and
43.70\%, $+0.57$ points over the standard residual and $+0.53$ points over FBT),
whereas FBT stays close to the standard residual (37.48\% vs.\ 37.45\%). Across both prefill modes, ALF
ranks first on EM and F1 and the fixed nearest-four variant ranks second.

\begin{keyfinding}
ALF lowers held-out evaluation loss relative to both the standard-residual backbone and FBT across all three scales
(MoE-1.6B, MoE-2.6B, and MoE-4.0B). On MoE-4.0B, these gains
carry over to both teacher-forced downstream accuracy across all four domains and true greedy autoregressive generation
across all 13 completion benchmarks.
\end{keyfinding}

\subsection{Extension to attention residuals}
\label{sec:attnres}

\begin{figure}[!ht]
  \centering
  \includegraphics{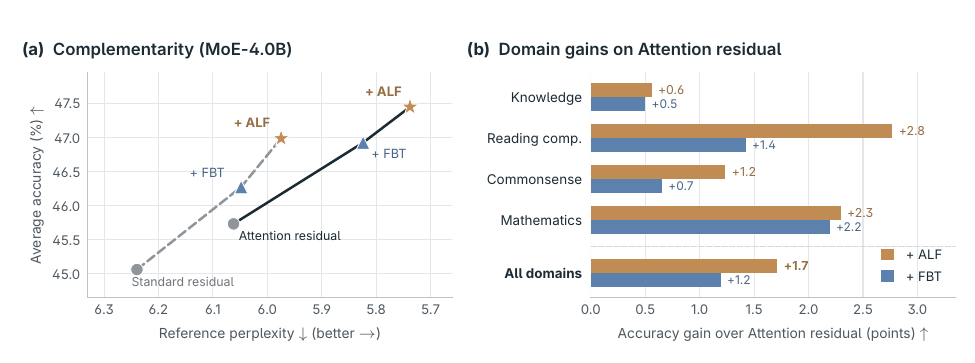}
  \caption[Downstream complementarity with Attention residual on MoE-4.0B.]{\textbf{Downstream complementarity with
  Attention residual on MoE-4.0B.} \textbf{(a)}~Four-domain average accuracy against reference-answer perplexity
  (reversed horizontal axis) for the Standard residual and Attention residual backbones, each with FBT and ALF
  (Tables~\ref{tab:downstream} and~\ref{tab:downstream_attnres}). \textbf{(b)}~Accuracy gains over the Attention
  residual backbone across the four domains and their average.}
  \label{fig:attnres_downstream}
\end{figure}

\paragraph{Complementarity with AttnRes.}
Because AttnRes replaces the standard residual's running sum with a learned within-position depth softmax
(\Cref{app:model}), we test whether cross-position feedback still helps on this stronger backbone by training full
(per-sublayer) AttnRes from scratch and branching ALF and FBT's input fusion under the same protocol
(Table~\ref{tab:main}). ALF lowers loss at all three scales by margins close to those on standard residuals and widens
its lead over FBT to at least 0.012 nats. On MoE-4.0B, ALF raises downstream average accuracy by
1.7 points and lowers reference-answer loss by 0.055 nats (Figure~\ref{fig:attnres_downstream};
per-task results in Table~\ref{tab:downstream_attnres}, \Cref{app:eval}), leading FBT in every domain, and an
independent backbone matches this gain (\Cref{app:setup}).

\begin{table}[t]
  \noindent
  \begin{minipage}[b]{.485\textwidth}
    \centering
    \footnotesize
    \setlength{\tabcolsep}{2.2pt}
    \renewcommand{\arraystretch}{1.55}
    \begin{tabular}{l rr @{\hspace{8pt}} rr}
\toprule
 & \multicolumn{2}{c}{\shortstack{Standard\\residual}} & \multicolumn{2}{c}{\shortstack{Attention\\residual}} \\
\cmidrule(lr){2-3}\cmidrule(l){4-5}
Model & Loss $\downarrow$ & Gain $\uparrow$ & Loss $\downarrow$ & Gain $\uparrow$ \\
\midrule
Backbone & 2.0577 & -- & 2.0424 & -- \\
+ Injection & 2.0388 & 0.0189 & 2.0312 & 0.0113 \\
+ Sublayer feedback & 2.0353 & 0.0224 & 2.0203 & 0.0222 \\
\midrule
\rowcolor{laOrangeLine!25}+ ALF (both) & \textbf{2.0333} & \textbf{0.0244} & \textbf{2.0193} & \textbf{0.0231} \\
\bottomrule
\end{tabular}
\par\vspace{0pt}
  \end{minipage}\hfill
  \begin{minipage}[b]{.485\textwidth}
    \centering
    \includegraphics[width=\linewidth]{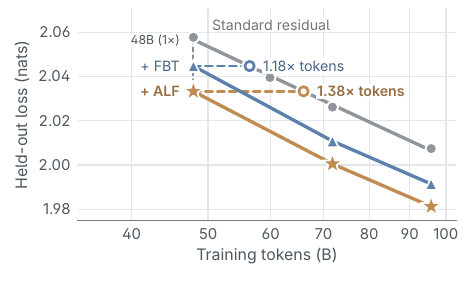}\par\vspace{0pt}
  \end{minipage}
  \par\noindent
  \begin{minipage}[t]{.485\textwidth}
    \vspace{0pt}
    \captionsetup{position=bottom,width=\linewidth}
\caption[Two paths across backbones.]{\textbf{Two paths across backbones.}
Held-out loss on MoE-1.6B in nats per token. Gains are relative to each backbone.
The single-path rows use that path alone, while ALF uses both. Bold marks the lowest loss and largest gain on each backbone.}
\label{tab:pathbackbones}

  \end{minipage}\hfill
  \begin{minipage}[t]{.485\textwidth}
    \vspace{0pt}
    \captionsetup{type=figure,position=bottom,width=\linewidth}
    \caption[Training data on MoE-1.6B.]{\textbf{Training data on MoE-1.6B.} The gray curve fits the
    standard model's loss from 48B tokens to twice that budget. Dashed lines mark where it
    reaches each feedback model trained on 48B tokens: 1.18$\times$ the budget for
    FBT and 1.38$\times$ for ALF.}
    \label{fig:tokens}
  \end{minipage}
\end{table}

\paragraph{Disentangling the two feedback paths.}
ALF combines two cross-position paths: additive token-input injection, a Recycler-style input-feedback baseline
\citep{bytedance2026recycler} rather than FBT's gated replacement, and per-sublayer feedback. To isolate what
drives ALF's complementarity with AttnRes and how sublayer feedback composes with input injection, we evaluate each
path alone on MoE-1.6B (Table~\ref{tab:pathbackbones}). Sublayer feedback alone achieves nearly identical gains on both
backbones, retaining 96\% of ALF's gain on AttnRes, whereas standalone token-input injection shrinks
to 49\%. Once AttnRes's within-position weighting already links every layer to the token input, per-sublayer
cross-position reads carry almost all of the remaining benefit. This explains why ALF widens its lead over FBT on
AttnRes while input injection still adds a small further gain when combined with sublayer feedback. Further integration
and map controls appear in \Cref{app:ablations}.

\begin{keyfinding}
ALF's gains in held-out loss and downstream accuracy carry over to the AttnRes backbone. Ablating the two paths confirms that both input injection and sublayer feedback contribute
to the gain, with sublayer feedback providing most of the benefit.
\end{keyfinding}

\section{Data and compute efficiency}
\label{sec:compute}

ALF reuses the previous position's computation while decoding each token in one pass. The maps and injection add
1.1\% to per-token decoding FLOPs on MoE-1.6B. During training, the second pass over the final quarter
raises total FLOPs to 1.253$\times$ the backbone (\Cref{app:cost}). We compare the loss reduction with this
training cost under two alternatives: adding training tokens at fixed model size, or increasing model size together with
its token budget.

At fixed model size, ALF uses training data more efficiently. We train the standard residual from scratch at
several token budgets on MoE-1.6B and fit a power law to held-out loss (Figure~\ref{fig:tokens})
\citep{wang2026full}. Matching ALF's loss requires 1.38$\times$ as many baseline training tokens,
equivalent to a 27\% reduction in tokens. The loss advantage persists when both models train
on twice the base token budget. This token advantage more than covers the extra training passes: dividing the
baseline's matched-loss token budget by ALF's training-cost multiplier gives an iso-loss compute ratio of
1.10$\times$ in favor of ALF. \citet{huang2026latentfeedback} also find that feedback improves
data efficiency when models train repeatedly on a fixed dataset.

\noindent
\begin{minipage}[c]{.485\textwidth}
\setlength{\parskip}{6pt}
When both model size and unique tokens can scale freely, spending the extra $25\%$ two-pass FLOPs on a larger
single-pass backbone achieves comparable pretraining compute efficiency; as \citet{huang2026latentfeedback} analyze,
latent feedback acts primarily as a data and fixed-size amplifier. Plotting each method against its total training FLOPs
across the three model sizes \citep{qin2026multi}, Figure~\ref{fig:scaling} compares ALF directly against FBT: accounting
for each method's exact two-pass overhead, FBT needs 1.07$\times$ the training FLOPs that ALF spends on
average to reach each ALF model's loss.

At decoding time, ALF runs in a single pass with only about 1.1\% extra forward FLOPs from the maps and input
projection, leaving per-token memory traffic (discussed in Limitations) as the main deployment tradeoff.
\end{minipage}\hfill
\begin{minipage}[c]{.485\textwidth}
  \centering
  \includegraphics[width=\linewidth]{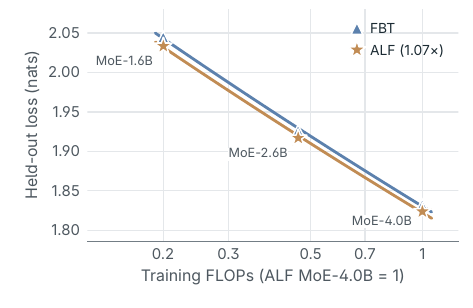}
  \captionsetup{type=figure,width=\linewidth,aboveskip=7pt}
  \caption[Held-out loss against training compute.]{\textbf{Held-out loss against training compute.}
  Training FLOPs are normalized to ALF on MoE-4.0B (including its second pass). Curves fit the power law
  $\mathcal{L}(C)=aC^{-\gamma}$ to measured losses (points); the legend ratio averages FBT's fitted FLOPs divided
  by ALF's at each ALF loss.}
  \label{fig:scaling}
\end{minipage}

\par

\begin{keyfinding}
At fixed size, ALF uses fewer tokens and less training compute to match the standard residual's loss, and across scales
it outperforms FBT in training compute efficiency.
\end{keyfinding}

\section{Analysis}
\label{sec:analysis}

We now examine which previous-position representations drive ALF's gains, how much of that gain survives when a fixed
nearest-four window replaces the learned depth softmax, how additional parallel passes behave beyond two-pass training,
and how prefill choices affect continuation quality.

\subsection{What the sublayer feedback reads}
\label{sec:maps}

\begin{figure}[!ht]
  \centering
  \includegraphics{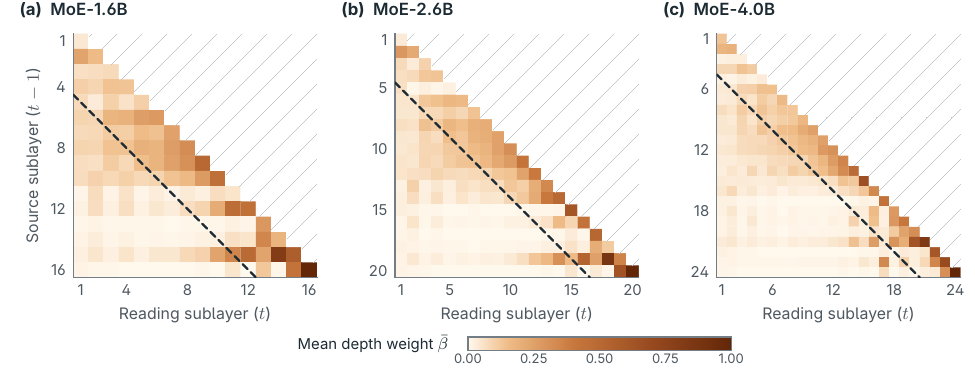}
  \caption[What the sublayer feedback reads.]{\textbf{What the sublayer feedback reads.} \textbf{(a--c)}~Learned
  attention maps over depth at the final checkpoints of MoE-1.6B, MoE-2.6B, and MoE-4.0B on the standard residual
  (\Cref{app:reads}). Each column gives one sublayer's mean weights over outputs of the previous position, and
  hatching marks sources outside its range. The dashed diagonal marks the lower boundary of each sublayer's current
  and next three deeper sublayers, where most of the attention weight concentrates.}
  \label{fig:reads}
\end{figure}

The learned attention maps in Figure~\ref{fig:reads}a--c show a consistent local diagonal pattern across all three model
sizes. Most of the attention probability mass falls on each sublayer's own depth and the next three deeper sublayers of
the previous position (above the dashed diagonal line), whereas more distant deeper outputs receive far less weight. At
all three sizes, every reader in the lower half gives more weight to the two nearest eligible outputs together than a
uniform read would: on MoE-4.0B, their combined weight averages 31\% across those readers, compared with
11\% under uniform weights. Across the network, shallow sublayers also carry the largest learned feedback
shares (the lower half accounts for 73\% of the summed shares on MoE-4.0B), so share-weighted reads focus on
nearby intermediate blocks of the previous position (\Cref{app:reads}).

Interventions on the trained model test how these reads contribute to prediction (\Cref{app:reads}). Keeping only the
most useful middle block retains 40\% of the sublayer feedback's contribution, while the top quarter of
blocks together retains 21\%. The observed preference for intermediate blocks is thus accompanied by a
larger contribution from those outputs.

\begin{table}[!ht]
  \centering
  \captionsetup{position=top}
  \caption[Feedback sources on MoE-1.6B.]{\textbf{Feedback sources on MoE-1.6B.}
  Held-out loss (nats per token, three passes) under Table~\ref{tab:main}'s standard-residual setup when
  varying which previous-position states each path reads. Within each group, $\Delta$ loss is relative to the
  first row (negative is better), lowest loss is bold, and the shaded row marks full ALF.}
  \label{tab:readsdepth}
  \footnotesize
  \setlength{\tabcolsep}{5.5pt}
  \begin{tabular}{llrr}
\toprule
\shortstack[l]{Feedback source\\(at every sublayer)} & \shortstack[l]{Injection source\\(at the token input)} & Loss $\downarrow$ & $\Delta$ loss \\
\midrule
\multicolumn{4}{l}{\textit{Feedback only: restrict its source}} \\
All eligible sublayer outputs & Disabled & \textbf{2.0353} & -- \\
Final state only & Disabled & 2.0414 & $+0.0061$ \\
\midrule
\multicolumn{4}{l}{\textit{Injection only: move the injection source}} \\
Disabled & Final state (block 8) & 2.0388 & -- \\
Disabled & State after block 6 & \textbf{2.0344} & $-0.0044$ \\
\midrule
\multicolumn{4}{l}{\textit{Both paths: restrict the feedback source}} \\
\rowcolor{laOrangeLine!25}All eligible sublayer outputs & Final state (block 8) & \textbf{2.0333} & -- \\
Last-block outputs only & Final state (block 8) & 2.0384 & $+0.0051$ \\
Nearest four outputs only & Final state (block 8) & 2.0346 & $+0.0013$ \\
\bottomrule
\end{tabular}

\end{table}

While those interventions zero out reads in an already trained model, the source controls in Table~\ref{tab:readsdepth}
retrain with restricted sources so the model can adapt to the depths available. Restricting sublayer feedback to only
the final state raises loss by 0.0061 nats without input injection, and restricting it to the last block
still raises loss by 0.0051 nats with both paths active. Replacing learned attention over depth with an equally
weighted read of the four nearest eligible outputs raises MoE-1.6B loss by only 0.0013 nats in
Table~\ref{tab:readsdepth}, which we examine across scales, downstream benchmarks, and read costs in \Cref{sec:window}.
As a post-hoc diagnostic guided by Figure~\ref{fig:reads}, moving standalone input injection from the last block down to
block six also lowers MoE-1.6B loss by 0.0044 nats, corroborating that intermediate layers carry the
strongest cross-position signal.

Unlike single-block input tapping, however, per-sublayer feedback discovers and routes these intermediate depths
automatically per sublayer and per token without oracle block selection, and it transfers reliably to AttnRes where
standalone input injection retains less than half of ALF's gain (Table~\ref{tab:pathbackbones}). Once sublayer feedback is active,
moving the input injection to block six adds little further gain (\Cref{app:ablations}), so the full model keeps the
default top-layer state for input injection. Weight maps, interventions and retrained controls thus all point to
intermediate layers as the primary driver of ALF's gains.

\begin{keyfinding}
Learned attention over depth places most of its probability mass on each sublayer's current and next three sublayers of
the previous position. Interventions and retrained controls confirm that these nearby outputs drive most of ALF's gain.
\end{keyfinding}

\subsection{Fixed nearest-window feedback}
\label{sec:window}

\begin{table}[!ht]
  \centering
  \captionsetup{position=top}
  \caption[Fixed nearest-window feedback across scales.]{\textbf{Fixed nearest-window feedback across scales.}
  Standard-residual models under Table~\ref{tab:main}'s protocol, comparing learned attention (+~ALF) with an
  equal-weight average of the current and next three sublayers (+~Fixed nearest-4). Left: three-pass held-out loss;
  right: accuracy and perplexity on Table~\ref{tab:downstream}'s seventeen tasks. Bottom rows show differences
  from ALF and FBT, with the best value per column in bold.}
  \label{tab:window}
  \footnotesize
  \setlength{\tabcolsep}{4pt}
  \begin{tabular}{l rrr ccc}
\toprule
 & \multicolumn{3}{c}{\textbf{Held-out loss (nats) $\downarrow$}} & \multicolumn{3}{c}{\textbf{Downstream Acc (\%) $\uparrow$ / PPL $\downarrow$}} \\
\cmidrule(lr){2-4}\cmidrule(l){5-7}
Model & MoE-1.6B & MoE-2.6B & MoE-4.0B & MoE-1.6B & MoE-2.6B & MoE-4.0B \\
\midrule
Standard residual & 2.0577 & 1.9400 & 1.8456 & 31.98 / 9.65 & 38.95 / 7.63 & 45.06 / 6.24 \\
+ FBT & 2.0447 & 1.9261 & 1.8322 & 32.47 / 9.47 & 40.21 / 7.32 & 46.27 / 6.05 \\
\rowcolor{laOrangeLine!25}+ ALF & \textbf{2.0333} & \textbf{1.9170} & \textbf{1.8238} & 33.29 / \textbf{9.22} & \textbf{40.56} / 7.22 & \textbf{46.98} / \textbf{5.97} \\
\rowcolor{laOrangeLine!12}+ Fixed nearest-4 & 2.0346 & 1.9188 & 1.8262 & \textbf{33.40} / \textbf{9.22} & 40.48 / \textbf{7.19} & 46.75 / 6.00 \\
\cmidrule(lr){1-7}
\textit{Fixed $-$ ALF} & $+0.0013$ & $+0.0018$ & $+0.0024$ & $+0.11$ / $+0.00$ & $-0.08$ / $-0.04$ & $-0.23$ / $+0.02$ \\
\textit{Fixed $-$ FBT} & $-0.0101$ & $-0.0072$ & $-0.0060$ & $+0.93$ / $-0.25$ & $+0.27$ / $-0.14$ & $+0.48$ / $-0.05$ \\
\bottomrule
\end{tabular}

\end{table}

\paragraph{Why a four-output window.}
The learned attention maps in Figure~\ref{fig:reads}a--c reveal a narrow diagonal band just above each sublayer's own
depth, an inductive bias discovered by training full cross-depth attention rather than prescribed in advance. Outputs
one to two blocks deeper at the previous position already carry refined context yet remain close to the reader's
representation scale, whereas top layers specialize toward the output head. Averaging across receiving sublayers
weighted by each sublayer's learned feedback share $\rho_l$ (\Cref{app:reads}), the four nearest eligible outputs (two
transformer blocks, clipped at the top) receive 73\%, 68\%, and 64\%
of the attention mass on MoE-1.6B, MoE-2.6B, and MoE-4.0B, compared with 46\%,
38\%, and 32\% under uniform weights. To separate the benefit of reading this local
band from per-token softmax selection, we replace \eqref{eq:hsoftmax} with an equal-weight average over those four
outputs while keeping the rank-64 maps, learned mixing shares, and token-input injection unchanged. Removing the softmax
drops all depth queries and reduces total cross-layer source reads per token from quadratic ($L(2L{+}1)$) to linear
($8L{-}6$) in depth, cutting reads by 57\% to 70\%.

\paragraph{What the architectural path carries.}
Across all three scales in Table~\ref{tab:window}, this fixed nearest-four window retains 94\% to
89\% of ALF's held-out loss reduction over the standard residual and 88\% to 72\%
of its margin over FBT. Downstream accuracy, autoregressive generation (Table~\ref{tab:generation}), and answer perplexity follow the same
pattern, tracking ALF closely while staying ahead of FBT at every scale. Adding evaluation passes beyond training also keeps lowering loss for the fixed
window ($-$0.0014 to $-$0.0012 nats from pass~2 to pass~4), matching ALF's pass-to-pass stability in
\Cref{sec:passes}, whereas FBT rebounds after pass~2. Direct per-sublayer access to nearby deeper outputs of the
previous position therefore drives both multi-pass stability and most of ALF's gain at these scales, even without
per-token softmax selection.

\paragraph{Where learned attention earns its margin.}
At the same time, the gap between the fixed window and learned ALF widens monotonically across all three scales: the
held-out loss penalty nearly doubles from 0.0013 nats on MoE-1.6B ($L=8$) to 0.0024 nats on MoE-4.0B
($L=12$), the share of ALF's margin over FBT lost by dropping the softmax grows from 12\% to
28\%, and downstream accuracy shifts monotonically from $+$0.11 to $-$0.23 points
relative to ALF. Depth geometry explains this divergence: at $L=8$ blocks, a four-sublayer window already spans
46\% of uniform share-weighted mass, leaving limited room to separate, whereas by $L=12$ uniform
coverage falls to 32\% and Figure~\ref{fig:reads} spreads weight across more layers (nearest-four mass
drops from 73\% to 64\% and mean top-1 weight from 0.41 to
0.37). ALF thus spans two operating points of one mechanism: the fixed nearest-four window offers a linear-read variant
discovered from ALF's attention maps at moderate depths, whereas learned attention recovers a growing share of the gain
as models deepen.

\begin{keyfinding}
Truncating the attention window to an equal-weight average of the current and next three sublayers preserves
most of the performance gain while cutting read cost from quadratic to linear in depth.
\end{keyfinding}

\subsection{Behavior beyond two training passes}
\label{sec:passes}

\begin{figure}[!ht]
  \centering
  \includegraphics{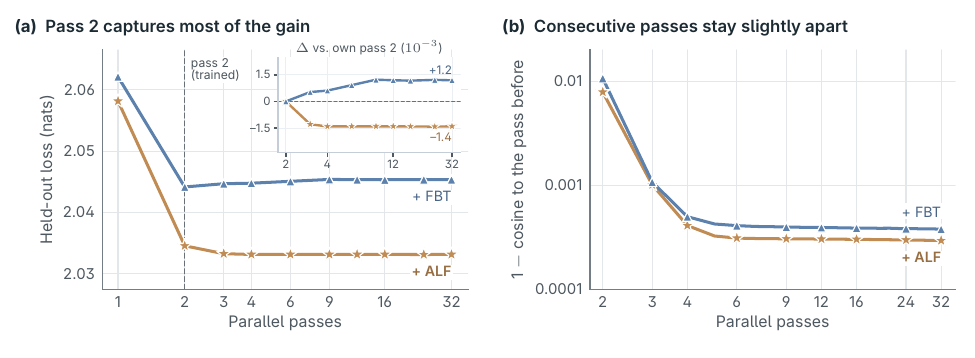}
  \caption[Behavior beyond the training passes on MoE-1.6B.]{\textbf{Behavior beyond the training passes on MoE-1.6B.}
  One evaluation runs 32 parallel passes over the held-out set for FBT and ALF, with the recipes of
  Table~\ref{tab:main}. \textbf{(a)}~Held-out loss from pass 1 (no feedback) through pass 32; the inset zooms
  in on each model's change relative to its own pass 2 (in $10^{-3}$ nats). \textbf{(b)}~One minus the cosine
  similarity between the final states of consecutive passes, starting at pass 2. Pass 2 captures most of the loss
  reduction and both losses settle within a few passes, while small differences between successive final states persist.}
  \label{fig:converge}
\end{figure}

The procedure in \Cref{sec:training} trains with two parallel passes, while decoding uses the states of
the completed previous token. Evaluating additional parallel passes lets feedback propagate farther along
the sequence and measures how the loss changes beyond the training regime
(Figure~\ref{fig:converge}). After the second pass, ALF's loss falls by a further 0.0014
nats and FBT's rises by 0.0012 nats. Both losses settle within a few passes.

State measurements confirm this stabilization: consecutive passes reach a final-state cosine similarity above
0.9996, while their relative state change levels off near 2\%, and longer evaluations reproduce the same
pattern (\Cref{app:passes}). Training with a third pass improves ALF's late-pass loss by only 0.0010 nats,
supporting the two-pass schedule at lower training cost (\Cref{app:passes}).

\subsection{Prompt processing and training choices}
\label{sec:recipe}

The prompt provides the states from which recurrent decoding begins. Building on the Fused and Soft prompt prefill
modes evaluated under greedy generation in Table~\ref{tab:generation}, Table~\ref{tab:prompt} isolates their effect on
teacher-forced continuation loss alongside a \emph{Standard} control that disables cross-position feedback throughout:
Fused applies feedback in both prompt and continuation, whereas Soft runs one plain backbone pass on the prompt before
enabling feedback on the continuation. This separates the benefit of feedback during continuation from the additional
benefit of processing the prompt with feedback.

\begin{table}[!ht]
  \centering
  \captionsetup{position=top}
  \caption[Prompt processing on MoE-1.6B.]{\textbf{Prompt processing on MoE-1.6B.}
  Continuation loss in nats after a 1,024-token prompt, using Table~\ref{tab:main}'s FBT and ALF recipes and
  a matched backbone control. Standard disables feedback throughout. Fused processes the prompt with nine
  feedback passes; Soft uses one plain pass. Both use nine feedback-enabled continuation passes to propagate feedback
  under teacher forcing. The backbone supports only Standard mode. Bold marks the lowest loss
  in each mode.}
  \label{tab:prompt}
  \small
  \setlength{\tabcolsep}{9pt}
  \begin{tabular}{l rrr r}
\toprule
 & \multicolumn{3}{c}{Continuation loss $\downarrow$} & Prefill penalty \\
\cmidrule(lr){2-4}\cmidrule(l){5-5}
Model & Standard & Fused & Soft & Soft $-$ Fused \\
\midrule
Standard backbone & \textbf{1.9660} & -- & -- & -- \\
+ FBT & 1.9704 & 1.9522 & 1.9621 & $+0.0099$ \\
\rowcolor{laOrangeLine!25}+ ALF & 1.9668 & \textbf{1.9405} & \textbf{1.9436} & $+0.0031$ \\
\bottomrule
\end{tabular}

\end{table}

Enabling feedback during continuation lowers loss for both methods, and using it during prompt processing
lowers loss further. Switching from Fused to Soft costs 0.003 nats for ALF and
0.010 nats for FBT. ALF therefore retains more of its benefit with a single plain prompt pass.
The Standard-mode controls show that these improvements depend on using feedback at evaluation time:
with feedback disabled, neither feedback-trained checkpoint improves on the standard backbone.

We also vary FBT's training modifications \citep{wang2026full}: uniform jitter on the fed-back state and a
prefix mixin that feeds plain embeddings on random document prefixes. Table~\ref{tab:recipe} (\Cref{app:passes}) shows
a tradeoff for FBT: keeping the prefix mixin raises held-out loss by 0.010 nats, whereas removing it leaves
FBT worse than the standard backbone in the alternating-turn evaluation, which repeatedly switches from plain prefill to
feedback-enabled continuation. ALF avoids this tradeoff. Without either the prefix mixin or jitter (which changes losses
by at most 0.0005 nats), it reaches the lowest loss across held-out evaluation, plain-prefill continuation
and alternating-turn continuation.

\FloatBarrier
\section{Conclusion}
\label{sec:conclusion}

Attention Latent Feedback (ALF) gives every sublayer direct access to deeper outputs of the previous position while
training in two parallel passes and decoding in one pass per token with minimal forward overhead. Across the
mixture-of-experts scales we study, ALF lowers held-out loss on both standard and attention residuals, outperforms FBT
under the same schedule, and improves both teacher-forced downstream accuracy and single-pass greedy autoregressive
generation. Depth weight maps, interventions and retrained source
controls all point to intermediate representations as the main gain beyond final-state feedback. At fixed model size,
ALF reaches a target loss with fewer training tokens, and most of its benefit persists after a single plain prompt pass.

\paragraph{Limitations.}
During decoding, each sequence caches the preceding token's $2L$ sublayer outputs ($O(Ld)$ memory, small relative to the
KV cache), yet reading them at every sublayer raises per-token memory traffic over final-state feedback. Speculative
decoding for final-state feedback uses stand-in states and correction passes \citep{huang2026feedbackshortcut}.
Extending it to per-sublayer reads remains open. Block-level
summaries as in Block AttnRes \citep{kimiteam2026attention} could reduce this traffic, while the fixed nearest-four
window of \Cref{sec:window} already removes 57\%--70\% of cross-layer reads with minimal loss
at the scales tested. Tapping an intermediate block for single-vector input injection (Table~\ref{tab:readsdepth}) is
also a lightweight alternative on standard residuals whose behavior at larger scales, on AttnRes, and under plain
prefill remains untested. We evaluate base MoE models up to 4.0B parameters and 128B tokens on short-answer and
multiple-choice tasks with limited seed repeats; scaling further and testing post-trained long-form generation remain
future work.

\newpage
\bibliographystyle{ims}
\bibliography{refs}

\newpage
\appendix
{\makeatletter\patchcmd{\l@section}{\addvspace{1.0em \@plus\p@}}{\addvspace{0.3em \@plus\p@}}{}{}\makeatother
 \color{sectionblue}\tableofcontents}
\section{Method details}
\label{app:arch}

\subsection{ALF on Attention Residuals}
\label{app:model}

\begin{figure}[t]
  \centering
  \includegraphics{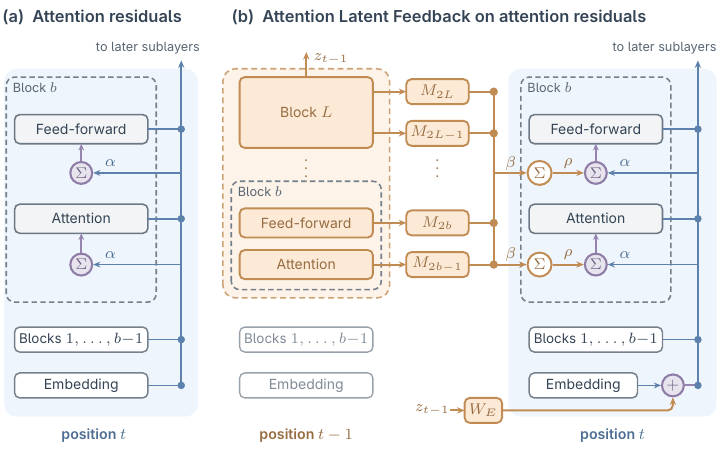}
  \caption{\textbf{Attention residuals, alone and with Attention Latent Feedback.} Blue marks the current stream,
  orange the sublayer feedback and its softmax, and purple the mixture $\Sigma$ that forms the input of each sublayer.
  \textbf{(a)}~In AttnRes, the current stream carries the sources of position $t$ upward, and each sublayer of block $b$
  takes a softmax over those below it, with weights $\walpha$. \textbf{(b)}~With ALF on AttnRes, every output of position
  $t-1$ passes through a per-output map, and each sublayer of block $b$ mixes a separate softmax over blocks $b$ to $L$,
  with weights $\wbeta$, into its current-stream softmax at share $\rho$. As in Figure~\ref{fig:arch}, the state
  $z_{t-1}$ also joins the token embedding through $W_E$ \eqref{eq:embed-inject}. Faded blocks of position $t-1$ are not read.}
  \label{fig:arch_attnres}
\end{figure}

\paragraph{Current-stream attention.}
The extension in \Cref{sec:attnres} uses full Attention Residuals \citep{kimiteam2026attention} as the backbone.
In \eqref{eq:xupdate}, each sublayer $l$ forms its \cstream{} $x_{t,l}$ using a learned pseudo-query
$w_l\in\mathbb{R}^d$ that scores all shallower sources at position $t$:
\begin{equation}
  x_{t,l}=\sum_{i=0}^{l-1}\walpha_{t,i\to l}\,v_{t,i},
  \qquad
  \walpha_{t,i\to l}=\frac{\exp\bigl(w_l^{\top}\RMSNorm(v_{t,i})/\sqrt{d}\bigr)}
                         {\sum_{i'=0}^{l-1}\exp\bigl(w_l^{\top}\RMSNorm(v_{t,i'})/\sqrt{d}\bigr)}.
  \label{eq:attnres}
\end{equation}
The output head uses a separate depth softmax over all $2L+1$ current-position sources to form the final state $z_t$
before final normalization. Pseudo-queries start at zero, giving uniform depth weights initially. ALF retains this
\cstream{} and mixes it with the \fbstream{} in \eqref{eq:mix} using $c_{t,l}=1$ \eqref{eq:scale}, since both streams
are already softmax-weighted averages.

\paragraph{Block-level construction.}
Figure~\ref{fig:arch_attnres} illustrates the full layout at block level. At the attention sublayer of block $b$, the
\fbstream{} starts at the previous position's attention output in that block; at the feed-forward sublayer, it starts
one sublayer higher because the current position's attention output is already available. Through \eqref{eq:embed-inject},
the projected final state $z_{t-1}$ enters $v_{t,0}$ and participates in every current-stream mixture, whereas the
output head reads only the \cstream{}.

\subsection{Parameterization, initialization, and input injection}
\label{app:params}
\label{app:injection}

\paragraph{Parameter count and initialization.}
Across all $2L$ sublayers, the low-rank source maps $M_i=I+U_iD_i$ (\Cref{sec:feedback}) add $4rdL$ parameters, the
depth queries $u_l$ and mixing logits $a_l$ add $2dL+2L$ parameters, and the token-input projection $W_E$ adds $d^2$
parameters. Initializing $U_i=0$, $u_l=0$, $a_l=\log(1/99)$ and $W_E=0$ starts every source map at the identity, sets
uniform depth weights $\wbeta_{t,i\to l}$ in \eqref{eq:hsoftmax}, assigns a $1\%$ initial feedback share
$\rho_l=\sigma(a_l)=0.01$ in \eqref{eq:mix}, and leaves the token embedding unmodified at the branch point.

\paragraph{Token-input injection.}
In \eqref{eq:embed-inject}, the embedding $e_t$ is taken after the backbone's embedding normalization, and the final
state $z_{t-1}$ is taken after the last block before final normalization ($x_{t-1,2L+1}$ on standard residuals, or the
pre-head depth mixture on AttnRes). The projection $W_E\in\mathbb{R}^{d\times d}$ is bias-free and initialized to zero,
and the $\RMSNorm$ applied to $z_{t-1}$ carries no learned gain. We train $W_E$ with Adam under zero weight decay and
scale its learning rate by $\kappa/d$ with $\kappa=7.68$ (giving factors of $0.01$, $0.0086$ and $0.0075$ across the
three MoE widths) to keep output updates width-invariant after branching, following the maximal-update parametrization
\citep{yang2021feature,yang2021tuning,yang2023spectral}.

\paragraph{Document boundaries and exact prefixes.}
Both cross-position pathways are disabled at document starts and throughout the first training pass, reducing pass~1 to
the plain backbone ($v_{t,0}=e_t$ and $h_{t,l}=x_{t,l}$). Under Jacobi unrolling (\Cref{sec:training}), pass~$k$ matches
the exact sequential recurrence on the first $k$ positions of each document: pass~1 is exact at position~1, and each
additional pass extends the exact prefix by one position. Full backpropagation through \eqref{eq:loss} lets later-pass
losses train the earlier-pass outputs they consume.

\subsection{Training and decoding compute}
\label{app:cost}
\label{app:compute}

\paragraph{FLOP accounting.}
We count matrix products at two FLOPs per active parameter and token in a forward pass, including the output head, and
add the score and value products over each attention layer's context, counting the backward pass as twice the forward
pass. Applying each low-rank map once adds $8rdL$ forward FLOPs per token, and the input projection adds $2d^2$.
Denoting their combined fraction of the backbone's forward FLOPs by $\mu$, running a second pass over the final quarter
of training (\Cref{sec:training}) gives the total training cost ratio
\begin{equation}
  \frac34+\frac14(2+\mu)=1+\frac{1+\mu}{4},
\end{equation}
which equals 1.253 on MoE-1.6B. Single-pass decoding incurs only the forward fraction $\mu$. Depth softmaxes are
omitted from these counts because they add less than $0.5\%$ to a forward pass.

\paragraph{Measured step times.}
For the full MoE-1.6B model on the standard residual, the median step times over the last 500 logged
training steps are 1.99 seconds for the continued backbone and 4.12 seconds for ALF. Both runs use
the same accelerator type, global batch and sequence length, with one microbatch for the backbone and four for the
two-pass model. The measured step-time ratio is 2.07$\times$; applying it to the two-pass final quarter gives
an estimated full-schedule ratio of 1.27$\times$, excluding compilation, evaluation and checkpoint I/O.

\FloatBarrier
\section{Experimental setup and evaluation}
\label{app:expsetup}

\subsection{Training setup}
\label{app:setup}

\urldef{\marinscalingladder}\url{https://github.com/marin-community/marin/blob/main/experiments/grug/moe_hero_ep/launch_scaling_ladder.py#L12-L23}
\paragraph{Model configurations and data.}
Table~\ref{tab:models} in \Cref{sec:setup} lists the three MoE configurations, taken specifically from Marin's
\texttt{moe\_hero\_ep} scaling ladder \citep{marin2026}\footnote{Defined at \marinscalingladder{}, with MoE-2.6B
interpolated between the MoE-1.6B and MoE-4.0B rungs.} rather than Marin's other MoE ablation suites. Every fourth
attention layer uses the full 4{,}096-token context, whereas the remaining layers use a 2{,}048-token sliding window. Query heads have width 128 and share one key-value head, or two on MoE-4.0B. Token budgets range from
48B to 128B. We use the Marin tokenizer \citep{marin2026}, with document boundaries masked in both
self-attention and cross-position feedback.

\begin{table}[t]
  \centering
  \captionsetup{position=top}
  \caption[Optimization settings.]{\textbf{Optimization settings.} Peak learning rates and size-dependent Adam
  hyperparameters from the exported training configurations. The feedback models inherit their backbone's schedule at
  the branch point. The input projection's additional learning-rate multiplier is given in \Cref{app:injection}.}
  \label{tab:optimizer}
  \small
  \setlength{\tabcolsep}{8pt}
  \begin{tabular}{l rrr}
\toprule
Parameter & MoE-1.6B & MoE-2.6B & MoE-4.0B \\
\midrule
Matrix learning rate & 0.0158 & 0.0148 & 0.0144 \\
Adam learning rate & 0.00365 & 0.00343 & 0.00333 \\
Adam $\beta_2$ & 0.96849 & 0.95598 & 0.95000 \\
Adam $\epsilon$ & $1.03\times10^{-15}$ & $1.13\times10^{-15}$ & $1.20\times10^{-15}$ \\
\bottomrule
\end{tabular}

\end{table}

\paragraph{Optimization and precision.}
The backbone follows Marin's matrix optimizer with momentum 0.95, while the remaining parameters use Adam with
$\beta_1=0.9$. Table~\ref{tab:optimizer} lists peak learning rates and size-dependent Adam settings. Learning
rates warm up over the first 1\% of training and decay linearly to 5\% of their peaks. Weight decay is
0.1, reduced to 0.02 for the router. New sublayer-feedback parameters use Adam, while the input
projection uses no weight decay and the inverse-width rate of \Cref{app:injection}. Parameters are stored in float32
and computation runs in bfloat16. Each main comparison uses one run per method and size, complemented by the
independent-backbone repeat below.

\paragraph{FBT baseline configuration.}
FBT replaces the token embedding with a learned gated fusion of the embedding and the preceding final state
\citep{wang2026full}, passed through the backbone's input normalization so that it enters at the embedding's scale.
Because omitting the prefix mixin yields substantially lower held-out loss (\Cref{sec:recipe}), we evaluate FBT without
the prefix mixin across all main comparisons; uniform jitter has negligible effect (at most 0.0005 nats in
Table~\ref{tab:recipe}, \Cref{app:passes}) and is retained on the standard-residual runs while omitted on AttnRes.

\paragraph{Evaluation passes and variance checks.}
All multi-pass MoE models use three evaluation passes on the six validation domains; a fourth pass shifts ALF's loss on
MoE-1.6B by only 0.0001 nats. Paired branches share their initialization checkpoint and data order. Repeating a
branch across two accelerator clusters changes held-out loss by at most 0.0001 nats, while independently trained
backbones differ by 0.011 nats. For single-scale ablation decisions, we use 0.003 nats as a practical reference threshold while still tracking
consistent monotonic trends across scales. Branching ALF from the second independent AttnRes backbone gains
0.022 nats, within 0.0008 nats of the gain on the first.

\begin{figure}[t]
  \centering
  \includegraphics{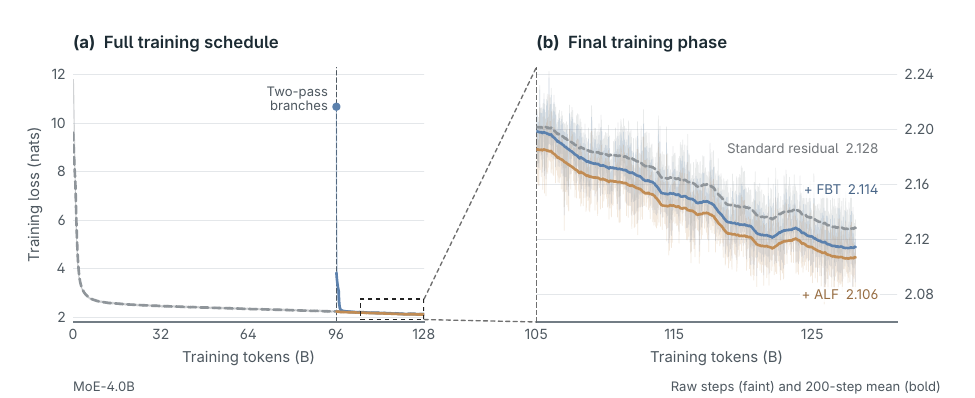}
  \caption[Training trajectories on MoE-4.0B.]{\textbf{Training trajectories on MoE-4.0B.}
  \textbf{(a)}~The full schedule, with FBT and ALF branching at $75\%$ of training.
  \textbf{(b)}~The final interval marked in (a), with a tighter loss scale. The dashed box is padded vertically for visibility. Faint lines show every logged step, and bold lines show a centered
  200-step mean with shorter windows at the endpoints. Labels report the mean of the final 200 steps.
  The plotted loss is the last-pass training cross-entropy, including the z-loss.}
  \label{fig:training}
\end{figure}

\paragraph{Training trajectories.}
Figure~\ref{fig:training} plots the full MoE-4.0B trajectories and enlarges the final phase. After branching at $75\%$
of training, ALF descends smoothly, whereas FBT shows an initial transient before both finish below the continued
backbone.
\FloatBarrier

\subsection{Downstream evaluation}
\label{app:eval}

\paragraph{Tasks and selection.}
The candidate suite combines the twenty tasks released by \citet{schwethelm2026recurrence} and
six tasks from DCLM CORE \citep{li2024datacomp}, counting HellaSwag once. Table~\ref{tab:evaltasks}
lists the selected tasks. The four domains follow \citet{saunshi2025reasoning}, including DROP under reading
comprehension because its arithmetic questions are grounded in a passage.

We retain tasks on which baseline accuracy is measurably above chance and improves during training. Specifically, on
MoE-1.6B with AttnRes, final accuracy must exceed both chance and accuracy at step 2{,}000 by at least
five binomial standard errors, and the standard-residual accuracy gain from MoE-1.6B to MoE-4.0B must exceed
ten times the larger within-task seed gap \citep{heineman2025signal}. This excludes SVAMP \citep{patel2021nlp}
and the seven symbolic probes (whose scores vary by 2.7 points between seeds). We also omit
tasks requiring long generated solutions or programs, including MATH500, HumanEval+, MBPP+ \citep{huang2026looped} and
BIG-Bench Hard with chain-of-thought \citep{suzgun2022challenging}, as our pretraining mixture contains 2.6\%
mathematics and 6.9\% code without instruction tuning.

\begin{table}[!ht]
  \centering
  \captionsetup{position=top}
  \caption[The downstream tasks.]{\textbf{The downstream tasks.} \emph{Shots} counts the solved examples in each prompt,
  \emph{Questions} the questions scored per model, and \emph{Chance} the accuracy of a uniform guess among the options.}
  \label{tab:evaltasks}
  \footnotesize
  \setlength{\tabcolsep}{3.5pt}
  \begin{tabular}{@{}ll l r r r@{}}
\toprule
Domain & Task & Format & Shots & Questions & Chance \\
\midrule
Knowledge & TriviaQA~\citep{joshi2017triviaqa} & Completion & 5 & 10{,}000 & -- \\
 & Natural Questions~\citep{kwiatkowski2019natural} & Completion & 5 & 3{,}610 & -- \\
 & WebQuestions~\citep{berant2013semantic} & Completion & 5 & 2{,}032 & -- \\
 & Jeopardy~\citep{li2024datacomp} & Completion & 10 & 2{,}117 & -- \\
 & QA-Wikidata~\citep{srivastava2022beyond} & Completion & 10 & 10{,}000 & -- \\
 & ARC-Easy~\citep{clark2018arc} & Multiple choice & 10 & 2{,}376 & 0.25 \\
 & ARC-Challenge~\citep{clark2018arc} & Multiple choice & 10 & 1{,}172 & 0.25 \\
\addlinespace[2pt]
Reading comprehension & LAMBADA~\citep{paperno2016lambada} & Completion & 0 & 5{,}153 & -- \\
 & SQuAD~\citep{rajpurkar2016squad} & Completion & 10 & 10{,}000 & -- \\
 & SQuAD 2.0~\citep{rajpurkar2018know} & Completion & 3 & 5{,}928 & -- \\
 & TyDi QA~\citep{clark2020tydi} & Completion & 3 & 440 & -- \\
 & CoQA~\citep{reddy2019coqa} & Completion & 1 & 7{,}983 & -- \\
 & DROP~\citep{dua2019drop} & Completion & 3 & 9{,}535 & -- \\
\addlinespace[2pt]
Commonsense reasoning & HellaSwag~\citep{zellers2019hellaswag} & Multiple choice & 0 & 10{,}000 & 0.25 \\
 & PIQA~\citep{bisk2020piqa} & Multiple choice & 10 & 1{,}834 & 0.50 \\
\addlinespace[2pt]
Mathematics & ASDiv~\citep{miao2020diverse} & Completion & 5 & 2{,}305 & -- \\
 & MAWPS~\citep{koncel2016mawps} & Completion & 5 & 1{,}772 & -- \\
\bottomrule
\end{tabular}

\end{table}

\paragraph{Prompts and scoring.}
Jeopardy, ARC-Challenge, SQuAD, HellaSwag and PIQA use DCLM CORE prompts and few-shot counts \citep{li2024datacomp},
while the remaining tasks use those of \citet{schwethelm2026recurrence}. For each of up to 10{,}000 questions
per task, we sample few-shot examples from the same task with a fixed seed, excluding the target question. Because all
tokens in each evaluation item are known in advance, FBT and ALF process the concatenated few-shot examples, question
and candidate answer together in three parallel Jacobi passes with feedback enabled throughout, and we score
only the answer positions on the final pass.

We follow the answer-scoring protocol of \citet{schwethelm2026recurrence}. For multiple-choice questions, we score each
option after the shared token prefix and select the option with the lowest mean negative log-likelihood per scored token
(omitting four PIQA questions where one tokenized option prefixes another, leaving no unshared tokens). For
completion tasks, we likewise use teacher forcing, conditioning each answer position on the prompt and preceding
reference tokens, and count a completion as correct only when the highest-probability next token matches the reference
at every answer position.

For each task, reference-answer perplexity is the exponential of the total negative log-likelihood over all scored
reference-answer tokens divided by their token count. Each domain reports the arithmetic mean of task accuracies and the
geometric mean of task perplexities, and the overall average weights all four domains equally.

\begin{table}[t]
  \centering
  \captionsetup{position=top}
  \caption[Downstream results on the Attention residual backbone.]{\textbf{Downstream results on the Attention residual backbone.}
  MoE-4.0B with AttnRes alone, with FBT and with Attention Latent Feedback (ALF), all at the final checkpoint (step 15{,}276).
  Accuracy and reference-answer perplexity are listed for the same seventeen tasks as Table~\ref{tab:downstream}.
  Each domain ends with its arithmetic mean accuracy and geometric mean perplexity.
  The final row weights the four domains equally; the best result for each task and average is in bold.}
  \label{tab:downstream_attnres}
  \footnotesize
  \setlength{\tabcolsep}{6pt}
  \begin{tabular}{@{}l ccc ccc@{}}
\toprule
 & \multicolumn{3}{c}{\textbf{Accuracy (\%) $\uparrow$}} & \multicolumn{3}{c}{\textbf{Perplexity $\downarrow$}} \\
\cmidrule(lr){2-4}\cmidrule(l){5-7}
Task & AttnRes & + FBT & + ALF & AttnRes & + FBT & + ALF \\
\midrule
\multicolumn{7}{@{}l}{\textit{Knowledge}} \\
TriviaQA & 22.3 & 22.7 & \textbf{23.2} & 12.00 & 11.60 & \textbf{11.37} \\
Natural Questions & 7.6 & 7.8 & \textbf{8.0} & 18.04 & 17.28 & \textbf{17.20} \\
WebQuestions & 13.3 & \textbf{14.7} & 14.5 & \textbf{14.97} & 15.13 & 15.03 \\
Jeopardy & 36.3 & 36.1 & \textbf{37.5} & 4.20 & 4.00 & \textbf{3.94} \\
QA-Wikidata & 64.3 & \textbf{65.9} & 65.6 & 3.87 & 3.57 & \textbf{3.55} \\
ARC-Easy & \textbf{73.7} & 72.9 & 72.4 & 7.78 & \textbf{7.53} & 7.65 \\
ARC-Challenge & 45.9 & \textbf{46.8} & 46.2 & 11.17 & \textbf{10.62} & 10.81 \\
\cmidrule(lr){1-7}
Average & 37.6 & 38.1 & \textbf{38.2} & 8.94 & 8.59 & \textbf{8.57} \\
\midrule
\multicolumn{7}{@{}l}{\textit{Reading comprehension}} \\
LAMBADA & 61.5 & 61.8 & \textbf{62.9} & 4.15 & 4.03 & \textbf{3.94} \\
SQuAD & 42.1 & 45.1 & \textbf{47.7} & 1.98 & 1.91 & \textbf{1.82} \\
SQuAD 2.0 & 34.6 & 38.7 & \textbf{40.4} & 2.27 & 2.14 & \textbf{2.05} \\
TyDi QA & 41.4 & 40.2 & \textbf{41.6} & 1.68 & 1.67 & \textbf{1.63} \\
CoQA & 33.1 & 33.6 & \textbf{34.9} & 3.53 & 3.42 & \textbf{3.39} \\
DROP & 14.0 & \textbf{15.8} & \textbf{15.8} & 6.99 & 6.53 & \textbf{6.46} \\
\cmidrule(lr){1-7}
Average & 37.8 & 39.2 & \textbf{40.5} & 3.03 & 2.92 & \textbf{2.84} \\
\midrule
\multicolumn{7}{@{}l}{\textit{Commonsense reasoning}} \\
HellaSwag & 60.8 & 61.6 & \textbf{62.2} & 11.76 & 11.57 & \textbf{11.51} \\
PIQA & 74.9 & 75.5 & \textbf{76.0} & 12.29 & 12.05 & \textbf{11.92} \\
\cmidrule(lr){1-7}
Average & 67.9 & 68.5 & \textbf{69.1} & 12.02 & 11.81 & \textbf{11.72} \\
\midrule
\multicolumn{7}{@{}l}{\textit{Mathematics}} \\
ASDiv & 39.0 & \textbf{41.3} & 40.8 & 4.55 & 4.28 & \textbf{4.13} \\
MAWPS & 40.3 & 42.4 & \textbf{43.1} & 3.77 & 3.53 & \textbf{3.49} \\
\cmidrule(lr){1-7}
Average & 39.6 & 41.8 & \textbf{41.9} & 4.15 & 3.89 & \textbf{3.80} \\
\midrule
\textbf{All domains} & 45.7 & 46.9 & \textbf{47.4} & 6.06 & 5.82 & \textbf{5.74} \\
\bottomrule
\end{tabular}

\end{table}

\paragraph{Checkpoints and per-task results.}
Table~\ref{tab:downstream} (and MoE-4.0B in Table~\ref{tab:window}) evaluates FBT, ALF and the fixed-window variant at
step 15{,}000 and the standard residual at its final step 15{,}276 (its step-15{,}000 checkpoint was not retained, giving
the baseline 276 additional training steps); MoE-1.6B and MoE-2.6B in Table~\ref{tab:window} use matched final
checkpoints for all models. Across Table~\ref{tab:downstream},
ALF improves accuracy on fifteen tasks and reference-answer perplexity on sixteen tasks
relative to the standard residual. The AttnRes extension evaluates MoE-4.0B at its final checkpoint and two earlier
checkpoints after branching; Table~\ref{tab:downstream_attnres} reports the final checkpoint for every task alongside
domain averages. Across these checkpoints, ALF improves average accuracy by 1.4 points and
reference-answer loss by 0.043 nats, with its accuracy gain growing from 0.3 points early
after branching to 1.7 at the final checkpoint.

\paragraph{Uncertainty.}
Models are compared on the same questions. We draw 1{,}000 paired bootstrap samples within each task
to obtain a 95\% interval for the difference in each domain average. We call a difference resolved
when this interval excludes zero and its magnitude also exceeds three times the larger gap from two
MoE-1.6B seed pairs. These seed gaps range from 0.11 to 0.74 accuracy points across
domains. We apply the same procedure to perplexity. The interval measures uncertainty across questions, and seed gaps check training variability.

\paragraph{Autoregressive generation evaluation.}
Table~\ref{tab:generation} evaluates true greedy autoregressive decoding ($\arg\max$, up to 32 generated tokens) on the
13 completion benchmarks from Table~\ref{tab:downstream} (spanning Knowledge, Reading Comprehension, and
Mathematics, excluding the four multiple-choice tasks: ARC-Easy, ARC-Challenge, HellaSwag, and PIQA), using the same
few-shot prompts, question pools, and checkpoint alignment (step~15{,}000 for feedback models, step~15{,}276 for the
standard residual). During decoding, each model maintains exact key--value caches, rolling causal short-convolution
buffers, and the previous token's feedback states, so every generated token runs in a single forward pass. Before full
evaluation, we verified incremental prefill-plus-decode logits against uncached full-sequence forward passes across all
models and prefill modes. Few-shot completions are truncated at EOS or the first newline (and at the first
whitespace-delimited word for 0-shot LAMBADA), then scored with SQuAD-style normalized exact match (plus numeric
equivalence on ASDiv and MAWPS) and normalized token F1.

\begin{table}[!ht]
  \centering
  \captionsetup{position=top}
  \caption[Per-task autoregressive generation on MoE-4.0B.]{\textbf{Per-task autoregressive generation on MoE-4.0B
  (Fused prefill).} Greedy normalized EM (in percent, higher is better) on each of the 13 completion tasks
  from Table~\ref{tab:downstream} using two-pass prompt prefill followed by single-pass decoding. Bold marks the best
  score in each row. ALF improves over the standard residual on all 13/13 tasks and over FBT on
  11/13 tasks.}
  \label{tab:generation_tasks}
  \footnotesize
  \setlength{\tabcolsep}{7pt}
  \begin{tabular}{l rrrr}
\toprule
Task & Standard residual & + FBT & + Fixed nearest-4 & + ALF \\
\midrule
\multicolumn{5}{l}{\textit{Knowledge}} \\
\quad TriviaQA & 29.95 & 29.51 & 30.50 & \textbf{30.52} \\
\quad Natural Questions & 8.14 & \textbf{8.50} & 7.92 & 8.39 \\
\quad WebQuestions & 13.39 & 13.34 & \textbf{13.83} & 13.44 \\
\quad Jeopardy & 46.15 & 47.38 & \textbf{48.98} & 48.61 \\
\quad QA-Wikidata & 64.20 & 63.58 & 64.35 & \textbf{64.36} \\
\quad \textit{Domain average} & 32.37 & 32.46 & \textbf{33.12} & 33.06 \\
\midrule
\multicolumn{5}{l}{\textit{Reading comprehension}} \\
\quad LAMBADA & 60.24 & 60.53 & 60.97 & \textbf{61.67} \\
\quad SQuAD & 46.92 & 48.68 & 50.03 & \textbf{50.30} \\
\quad SQuAD 2.0 & 41.62 & 43.12 & \textbf{45.34} & 45.12 \\
\quad TyDi QA & 38.64 & 43.41 & \textbf{43.64} & 43.41 \\
\quad CoQA & 40.96 & 41.46 & 42.06 & \textbf{43.14} \\
\quad DROP & 15.20 & 15.42 & 16.33 & \textbf{16.46} \\
\quad \textit{Domain average} & 40.59 & 42.10 & 43.06 & \textbf{43.35} \\
\midrule
\multicolumn{5}{l}{\textit{Mathematics}} \\
\quad ASDiv & 38.35 & 40.74 & 40.91 & \textbf{42.30} \\
\quad MAWPS & 40.41 & 41.14 & 42.83 & \textbf{43.23} \\
\quad \textit{Domain average} & 39.38 & 40.94 & 41.87 & \textbf{42.76} \\
\midrule
\textbf{Domain-balanced average} & 37.45 & 38.50 & 39.35 & \textbf{39.73} \\
\textit{13-task macro average} & 37.24 & 38.22 & 39.05 & \textbf{39.30} \\
\bottomrule
\end{tabular}

\end{table}

Table~\ref{tab:generation_tasks} breaks down Fused-prefill EM across all 13 tasks for the four main MoE-4.0B
models: ALF improves over the standard residual on every task (13/13) and leads FBT on
11/13 tasks, with the largest margins on reading comprehension and mathematics (such as SQuAD,
SQuAD~2.0, CoQA, ASDiv, and MAWPS). On MoE-1.6B (step~11{,}420, Fused domain-balanced EM), both ALF (22.56\%)
and fixed nearest-four (22.61\%) likewise surpass FBT (20.87\%) and the standard residual
(20.31\%), mirroring the depth-scaling trend of \Cref{sec:window}.

\FloatBarrier
\section{Supplementary analysis}
\label{app:supplementary}

\subsection{Depth weights and interventions}
\label{app:reads}

\paragraph{Average depth weights.}
Panels~(a--c) of Figure~\ref{fig:reads} show the final-checkpoint softmax weights of \eqref{eq:hsoftmax} on the standard
residual averaged over a held-out set $\mathcal{T}$ of positions that read a preceding position, evaluated in pass~2
reading pass~1 (we omit pass superscripts below for readability):
\begin{equation}
  {\color{laOrangeText}\bar{\beta}_{i\to l}}=\frac{1}{|\mathcal{T}|}\sum_{t\in\mathcal{T}}\wbeta_{t,i\to l}.
\end{equation}
Each column sums to one over eligible sources $i\ge l$; the deepest reader has only one eligible source, so its weight
is one by construction. For the two nearest sources of a reader in the lower half, the uniform reference is
$2/(2L-l+1)$, and we average both this reference and the observed sum equally across those readers.

\paragraph{Share-weighted source mass.}
Because the \fbstream{} enters sublayer $l$ at share $\rho_l$, rescaled by $c_{t,l}$ \eqref{eq:scale} to match the
norm of the \cstream{}, we summarize source usage by the share-weighted average across receiving sublayers. The mass
assigned to source $i$ is
\begin{equation}
  r_i=\frac{\sum_{l=1}^{i}\rho_l\,{\color{laOrangeText}\bar{\beta}_{i\to l}}}{\sum_{l=1}^{2L}\rho_l},
  \label{eq:readshare}
\end{equation}
and block $b$ receives $R_b=r_{2b-1}+r_{2b}$. Because each sublayer's weights sum to one, the $r_i$ sum to one as well,
and $\sum_i r_i\,i/(2L)$ gives the mean source depth.

Because the inner sum in \eqref{eq:readshare} runs over sublayers $1$ to $i$ that can read source $i$, a deep output
collects weight from more readers than a shallow one even under uniform attention. To separate this geometric effect
from learned preferences, we hold $\rho_l$ fixed and replace ${\color{laOrangeText}\bar{\beta}_{i\to l}}$ with the
uniform weight $1/(2L-l+1)$ over the $2L-l+1$ eligible sources,
\begin{equation}
  r^{\mathrm{even}}_i=\frac{\sum_{l=1}^{i}\rho_l/(2L-l+1)}{\sum_{l=1}^{2L}\rho_l}.
\end{equation}
On MoE-4.0B, the learned mean depth is 0.55, compared with 0.70 under uniform weights. At all
three scales, blocks in the lower half receive more mass than this reference, whereas those in the top quarter receive
less. Using the same share weighting across receiving sublayers, the mass assigned to each sublayer's four nearest
eligible outputs in \Cref{sec:window} is
\begin{equation}
  m_{4}=\frac{\sum_{l=1}^{2L}\rho_l\sum_{i=l}^{\min(l+3,\,2L)}{\color{laOrangeText}\bar{\beta}_{i\to l}}}{\sum_{l=1}^{2L}\rho_l},
  \qquad
  m^{\mathrm{even}}_{4}=\frac{\sum_{l=1}^{2L}\rho_l\,\min(4,\,2L-l+1)/(2L-l+1)}{\sum_{l=1}^{2L}\rho_l},
\end{equation}
giving 73\%, 68\%, and 64\% against uniform references of
46\%, 38\%, and 32\% across MoE-1.6B, MoE-2.6B, and MoE-4.0B.

\paragraph{Single-block interventions.}
To isolate each block's causal contribution, we retain only the terms of block $b$ at every sublayer $l$,
\begin{equation}
  n_{t,l;b}=\sum_{i\in\{2b-1,\,2b\},\;i\ge l}\wbeta_{t,i\to l}\,M_i\,v_{t-1,i},
\end{equation}
and mix $c_{t,l}\,n_{t,l;b}$ into the sublayer input \eqref{eq:mix}, keeping the softmax weights
$\wbeta_{t,i\to l}$ and rescaling scalar $c_{t,l}$ \eqref{eq:scale} from the full read $n_{t,l}$ while leaving the
token-input injection \eqref{eq:embed-inject} active. Letting $\ell_{\mathrm{off}}$ denote the pass-2 held-out loss
with the \fbstream{} zeroed out, $\ell_{\mathrm{full}}$ the full pass-2 loss, and $\ell_b$ the loss under this
intervention, we report
\begin{equation}
  G_b=\frac{\ell_{\mathrm{off}}-\ell_b}{\ell_{\mathrm{off}}-\ell_{\mathrm{full}}},
\end{equation}
where one matches the full sublayer-feedback gain and zero matches removing it. Keeping the original weights and scale
measures each block's retained contribution in the trained mixture.
\FloatBarrier

\subsection{Additional ablation comparisons}
\label{app:ablations}

\paragraph{Overlap between the two paths.}
Combining both paths consistently yields the lowest held-out loss across all three model sizes and both backbones.
Adding token-input injection on top of sublayer feedback further lowers loss by 0.0020, 0.0015 and 0.0011
nats on MoE-1.6B, MoE-2.6B and MoE-4.0B on standard residuals, and by 0.0010, 0.0006 and
0.0002 nats on AttnRes. While each individual increment lies within our single-scale 0.003-nat threshold
(\Cref{sec:setup}), confirming that sublayer feedback carries the bulk of ALF's gain, the consistent directional
improvement across all six settings supports pairing both paths. Changing the injection's source to an intermediate
block helps when it is used alone (Table~\ref{tab:readsdepth}), whereas with both paths active on MoE-1.6B it improves
loss by only 0.0020 nats.

\begin{table}[!ht]
\centering
\captionsetup{position=top,width=\linewidth}
\caption[Feedback integration on MoE-1.6B.]{\textbf{Feedback integration on MoE-1.6B.}
Standard-residual models share Table~\ref{tab:main}'s feedback training schedule.
Loss is in nats per token after three evaluation passes.
Each pair's first row is its reference; $\Delta$ loss subtracts it from the second row. Lower loss is bold.}
\label{tab:ablate}
\begin{minipage}{0.8\textwidth}
  \centering
  \small
  \setlength{\tabcolsep}{6pt}
  \begin{tabularx}{\linewidth}{>{\raggedright\arraybackslash}Xrr}
\toprule
Configuration & Loss $\downarrow$ & $\Delta$ loss \\
\midrule
\multicolumn{3}{l}{\textit{Final-state feedback only (at the token input)}} \\
Additive injection only & \textbf{2.0388} & -- \\
FBT's gated replacement & 2.0447 & $+0.0059$ \\
\addlinespace[5pt]
\multicolumn{3}{l}{\textit{Whole-depth feedback only}} \\
Sublayer feedback & \textbf{2.0353} & -- \\
FBT's gated replacement at the token input & 2.0509 & $+0.0156$ \\
\addlinespace[5pt]
\multicolumn{3}{l}{\textit{Both paths}} \\
\rowcolor{laOrangeLine!25}ALF & \textbf{2.0333} & -- \\
ALF without maps & 2.0367 & $+0.0034$ \\
\bottomrule
\end{tabularx}

\end{minipage}
\end{table}

\paragraph{Input fusion and maps.}
Table~\ref{tab:ablate} compares feedback integration with the source held fixed within each pair.
At the token input, additive injection of the final state lowers loss by 0.0059 nats relative to
FBT's gated replacement. With both ALF paths active, removing the per-output maps raises loss by
0.0034 nats.

\paragraph{Source depth and integration.}
Combining the integration controls in Table~\ref{tab:ablate} with the final-state sublayer-mixture control in
Table~\ref{tab:readsdepth} yields four cells crossing the source (full depth or final state) with its entry path (the
sublayer mixture or FBT's gated input fusion), without the additional input injection. Using the gated input path
raises loss by 0.0156 nats for the full-depth source and 0.0033 nats for the final state.
These comparisons vary the entry point and fusion rule together.
Within the sublayer mixture, full depth improves loss by 0.0061 nats; within FBT's fusion it worsens
loss by 0.0062 nats. The resulting interaction is 0.0123 nats, showing that the value of a source
depends on how it is incorporated.

Pass-1 held-out losses (with feedback disabled) also differ. The two models using gated input replacement end
0.0044 nats (final-state FBT) and 0.0043 nats (full-depth gated replacement) above the continued
backbone, while the additive variants stay within 0.0008 nats. For full-depth input fusion, loss rises by
0.0170 nats from the second to the fourth evaluation pass. These measurements accompany the end losses in
Tables~\ref{tab:readsdepth} and~\ref{tab:ablate}.
\FloatBarrier

\subsection{Evaluation and training passes}
\label{app:passes}

\paragraph{Evaluation beyond the training passes.}
Figure~\ref{fig:converge} evaluates 32 passes for the MoE-1.6B recipes in Table~\ref{tab:recipe}.
Repeating the evaluation with 64 passes gives the same loss and state-change pattern: the mean relative
change in final-state norm remains near 2\% even after the held-out loss stabilizes. We use this pass-32
held-out loss from our standalone multi-pass evaluation harness (whose single-pass standard-residual baseline differs
from Table~\ref{tab:main} by $0.0003$ nats due to evaluation batching) to approximate recurrent state propagation under
teacher forcing in Tables~\ref{tab:passes} and~\ref{tab:recipe}; finite-pass evaluation is exact only on the prefix of
\Cref{app:params}.

\paragraph{Adding a training pass.}
\begin{table}[t]
  \centering
  \captionsetup{position=top}
  \caption[Two and three training passes on MoE-1.6B.]{\textbf{Two and three training passes on MoE-1.6B.} FBT and ALF
  on the standard residual with $K=2$ or $K=3$ passes in the last quarter, against the standard residual trained from
  scratch on 1.25 and 1.5 times the tokens. \emph{Compute} is relative to the backbone on the base budget,
  \emph{Pass~$K$} gives the held-out loss at the last training pass, and \emph{Pass~32} gives the converged
  held-out loss under teacher forcing. A third pass lowers ALF's pass-32 held-out loss by
  0.0010 nats, below the 0.003-nat threshold of \Cref{sec:setup}, so two passes suffice.}
  \label{tab:passes}
  \footnotesize
  \setlength{\tabcolsep}{6pt}
  \begin{tabular}{l c c ccc}
\toprule
Model & $K$ & Compute & Pass $K$ & Pass 32 & Pass 32 $-$ Pass $K$ \\
\midrule
Standard residual, $1.25\times$ tokens & 1 & 1.25 & 2.0395 & 2.0395 & -- \\
Standard residual, $1.5\times$ tokens & 1 & 1.50 & 2.0262 & 2.0262 & -- \\
\midrule
\quad + FBT & 2 & 1.25 & 2.0442 & 2.0454 & $+0.0012$ \\
\quad + FBT & 3 & 1.50 & 2.0426 & 2.0426 & $0.0000$ \\
\midrule
\rowcolor{laOrangeLine!25}\quad + ALF & 2 & 1.25 & 2.0346 & 2.0332 & $-0.0014$ \\
\quad + ALF & 3 & 1.51 & 2.0324 & 2.0322 & $-0.0001$ \\
\bottomrule
\end{tabular}

\end{table}

Table~\ref{tab:passes} compares two and three training passes from the same MoE-1.6B checkpoint, with matched
data and the main recipes. We set two criteria before these runs: the pass-32 held-out loss gain must exceed
0.003 nats, and the resulting ALF loss must improve on the standard residual trained with the same
compute spent on additional tokens.

The third pass lowers ALF's pass-32 held-out loss by 0.0010 nats, below the threshold, while the same
extra compute lowers the standard residual's loss by 0.013 nats. FBT gains 0.0027 nats from the third
pass, also below the threshold, although it reduces the gap between pass~$K$ and pass~32.
ALF's pass-32 held-out loss is already no higher than its pass-$K$ loss with either setting. We therefore
use two passes for both methods in the main comparison.

\paragraph{Prompt processing and training recipes.}
Table~\ref{tab:recipe} expands the main-checkpoint comparison in Table~\ref{tab:prompt} to all combinations
of jitter and prefix mixin. Prompt losses score the same continuation tokens in every mode, using nine
continuation passes for both Fused and Soft. The alternating-turn evaluation splits documents into
512-token turns, prefills even turns plainly, and scores odd turns with feedback enabled.
The held-out columns compare the last training pass with pass 32.

\begin{table}[!ht]
  \centering
  \captionsetup{position=top}
  \caption[Training recipe and prompt processing on MoE-1.6B.]{\textbf{Training recipe and prompt processing on MoE-1.6B.}
  Standard-residual MoE-1.6B models with FBT or ALF, trained with and without FBT's two stabilization aids
  \citep{wang2026full}, uniform jitter on the fed-back state and the prefix mixin. Entries are held-out losses in nats
  or differences of two: on the \emph{held-out set} at pass~2 and at pass~32 (the converged late-pass
  teacher-forced held-out loss, approximating recurrent state propagation without self-rollout); \emph{after a
  1,024-token prompt} with nine continuation passes, following either nine feedback-enabled prompt passes (Fused) or one
  plain prompt pass (Soft); and over the feedback-enabled odd turns of 512-token \emph{multi-turn} documents.
  Bold marks the lowest loss in each column except pass~2, and every loss within 0.0002 nats of it, and
  $\dagger$ marks the recipes of Table~\ref{tab:main}.}
  \label{tab:recipe}
  \footnotesize
  \setlength{\tabcolsep}{5pt}
  \begin{tabular}{l cc ccc ccc c}
\toprule
 & & & \multicolumn{3}{c}{Held-out set} & \multicolumn{3}{c}{After a 1,024-token prompt} & Multi-turn \\
\cmidrule(lr){4-6}\cmidrule(lr){7-9}\cmidrule(l){10-10}
Model & Jitter & \begin{tabular}[b]{@{}c@{}}Prefix\\mixin\end{tabular} & \begin{tabular}[b]{@{}c@{}}Pass 2\end{tabular} & \begin{tabular}[b]{@{}c@{}}Pass 32\end{tabular} & \begin{tabular}[b]{@{}c@{}}Pass 32\\$-$ Pass 2\end{tabular} & \begin{tabular}[b]{@{}c@{}}Fused\end{tabular} & \begin{tabular}[b]{@{}c@{}}Soft\end{tabular} & \begin{tabular}[b]{@{}c@{}}Soft\\$-$ Fused\end{tabular} & \begin{tabular}[b]{@{}c@{}}Scored\\turns\end{tabular} \\
\midrule
Standard residual &  &  & 2.0574 & 2.0574 & -- & 1.9660 & 1.9660 & -- & 1.9887 \\
\midrule
+ FBT & \checkmark & \checkmark & 2.0541 & 2.0549 & $+0.0008$ & 1.9602 & 1.9601 & $-0.0001$ & 1.9839 \\
+ FBT &  & \checkmark & 2.0541 & 2.0548 & $+0.0008$ & 1.9602 & 1.9601 & $-0.0001$ & 1.9838 \\
+ FBT$^\dagger$ & \checkmark &  & 2.0442 & 2.0454 & $+0.0012$ & 1.9522 & 1.9621 & $+0.0099$ & 1.9902 \\
+ FBT &  &  & 2.0442 & 2.0455 & $+0.0012$ & 1.9524 & 1.9625 & $+0.0101$ & 1.9906 \\
\midrule
+ ALF & \checkmark & \checkmark & 2.0418 & 2.0411 & $-0.0007$ & 1.9477 & 1.9488 & $+0.0011$ & 1.9732 \\
+ ALF &  & \checkmark & 2.0418 & 2.0411 & $-0.0007$ & 1.9477 & 1.9488 & $+0.0011$ & 1.9732 \\
+ ALF & \checkmark &  & 2.0346 & \textbf{2.0332} & $-0.0014$ & \textbf{1.9406} & \textbf{1.9436} & $+0.0031$ & \textbf{1.9692} \\
\rowcolor{laOrangeLine!25}+ ALF$^\dagger$ &  &  & 2.0346 & \textbf{2.0332} & $-0.0014$ & \textbf{1.9405} & \textbf{1.9436} & $+0.0031$ & \textbf{1.9693} \\
\bottomrule
\end{tabular}

\end{table}

\end{document}